\documentclass[conference]{IEEEtran}
\IEEEoverridecommandlockouts
\usepackage{cite}
\usepackage{amsmath,amssymb,amsfonts}
\usepackage{algorithmic}
\usepackage{graphicx}
\usepackage{textcomp}
\usepackage{hyperref}
\usepackage[]{algorithm2e}
\SetKwRepeat{Do}{do}{while}%

\usepackage[table,xcdraw]{xcolor}
\def\BibTeX{{\rm B\kern-.05em{\sc i\kern-.025em b}\kern-.08em
    T\kern-.1667em\lower.7ex\hbox{E}\kern-.125emX}}

\begin{document}

\IEEEoverridecommandlockouts

\IEEEpubid{\makebox[\columnwidth]{979-8-3503-1458-8/23/\$31.00~\copyright2023 IEEE \hfill} 
\hspace{\columnsep}\makebox[\columnwidth]{ }}

\title{Directed differential equation discovery using modified mutation and cross-over operators\\
}

\author{\IEEEauthorblockN{Elizaveta Ivanchik}
\IEEEauthorblockA{\textit{NSS Lab} \\
\textit{ITMO University}\\
Saint-Petersburg, Russia \\
eaivanchik@itmo.ru}
\and
\IEEEauthorblockN{Alexander Hvatov}
\IEEEauthorblockA{\textit{NSS Lab} \\
\textit{ITMO University}\\
Saint-Petersburg, Russia  \\
alex\_hvatov@itmo.ru}
}


\maketitle

\IEEEpubidadjcol

\begin{abstract}
The discovery of equations with knowledge of the process origin is a tempting prospect. However, most equation discovery tools rely on gradient methods, which offer limited control over parameters. An alternative approach is the evolutionary equation discovery, which allows modification of almost every optimization stage. In this paper, we examine the modifications that can be introduced into the evolutionary operators of the equation discovery algorithm, taking inspiration from directed evolution techniques employed in fields such as chemistry and biology. The resulting approach, dubbed directed equation discovery, demonstrates a greater ability to converge towards accurate solutions than the conventional method. To support our findings, we present experiments based on Burgers', wave, and Korteweg--de Vries equations. 
\end{abstract}


\section{Introduction}

The discovery of equations through machine learning has become an exciting area of research, building on the pioneering work of SINDy \cite{brunton2016discovering} and PDE-FIND \cite{rudy2017data} to generate concise and interpretable models for physical data. In this pursuit, a variety of techniques have been developed, including SINDy add-ons such as weak \cite{messenger2021weak} and ensemble methods \cite{fasel2022ensemble}, as well as other frameworks like PDE-Net \cite{long2019pde}, DLGA \cite{xu2020dlga}. However, a challenge of these symbolic regression methods is the reliance on a priori assumptions about which terms to include in the equations.

One may use different tools, including evolutionary approaches \cite{chen2022symbolic}, as well as other optimization methods \cite{lu2021bayesian}, to expand the variability of equation forms. Evolutionary optimization-based methods are particularly noteworthy as they do not rely on specific assumptions \cite{atkinson2019data}. Nonetheless, extensive exploration of the search space may yield non-physical model outcomes when dealing with experimental data.

In the pursuit of equation discovery, utilizing the known information regarding the underlying process can be alluring. Gradient methods offer the possibility of achieving this by altering the target function or imposing limitations, but this can result in a less stable optimization process. On the contrary, evolutionary methods offer a more seamless incorporation of knowledge.

A multitude of techniques exist for integrating knowledge into the general evolutionary algorithm. Some approaches involve leveraging non-random initial guesses generated from an expert solution \cite{hitomi2018incorporating}, while others rely on modifications to sub-optimal solution evaluation \cite{cuevas2020recent,martens2010automatically,li2022two,sarafanov2022evolutionary}. Intelligent data preprocessing methods can also be utilized to handle systematic observation errors and reduce the impact of noise \cite{zhang2022parsimony}. Another approach involves modifying the crossover and mutation operators to incorporate knowledge \cite{mahbub2016incorporating}. Additionally, knowledge-based hyper-parameter tuning is becoming increasingly popular \cite{wang2021experiencethinking}.

However, in the context of equation discovery, expert knowledge is often overlooked. The paper outlines a novel approach to incorporate domain knowledge into the equation discovery algorithm by employing modified mutation and cross-over strategies. The inspiration behind these changes is drawn from directed biological evolution \cite{wang2021directed}. Specifically, our methodology introduces the concept of ``gene importance'', which applies to the evolutionary optimization operators. Our discovery process is referred to as ``directed'' due to the inclusion of this concept. The addition of direction to the search enables us to converge to the correct equation more often and handle data and equations with that traditional equation discovery methods unable to work.

The paper compares the effectiveness of the directed approach to the classical approach, which is extensively described in \cite{maslyaev2021partial}, as well as the state-of-the-art equation discovery pySINDy \cite{desilva2020}. Our findings indicate that the proposed approach surpasses the classical evolutionary discovery method in both stability and the quality of the derived equations. We also demonstrate that, in some cases, our approach succeeds in producing an equation that the gradient-based pySINDy fails to identify. Although our method requires more optimization time, it often leads to the correct equation.

It should be noted that our study does not address the issue of knowledge extraction from data without any prior assumptions about the equation form. This aspect warrants further investigation. However, we do focus on describing how such knowledge can be integrated into the equation discovery process. Furthermore, if the methods of knowledge extraction and incorporation are used in conjunction, it could lead to the development of equation meta-learning, which is a natural extension of the equation discovery process

The rest of the paper is organized as follows. Sect.~\ref{sec:classics} describes the classical evolutionary equation discovery, Sect.~\ref{sec:modified_operators} describes what changes are made to the classical algorithm, Sect.~\ref{sec:experiments} shows the analysis of experiments on three model examples (Burgers' equation, wave equation and Korteweg--de Vries equation), Sect.~\ref{sec:conclusion} spotlights the main findings and proposes a direction for further work.

\section{Background}

In an ideal scenario, the discovery of differential equations would enable us to derive the complete underlying equations based on the observational data. Unfortunately, in practical situations, we can only approximate the system and obtain a rough estimation.

The discovery process involves data-driven structural optimization of the equation. We may choose two strategies for optimization:

\begin{itemize}
    \item The application of gradient optimization to a pre-determined form ($u_t=F(u,u_x,u_{xx},...)$) has been studied in previous works \cite{messenger2021weak,long2019pde,Kaptanoglu2022,desilva2020}. A notable benefit of this methodology is its high convergence speed. However, there are also some drawbacks, such as a fixed form and limited control over the optimization process. Typically, gradient optimization can only be influenced by choosing an appropriate initial guess. 
    \item{The  application of the evolutionary optimization to a model consists of small parametrized blocks. We do not make any a priori assumptions about the equation form, aside from the maximal order and length. Paper \cite{maslyaev2021partial} provides an example of this approach. While we are able to direct the evolution by including various heuristics, the model requires an extended optimization time, which is a disadvantage.}
\end{itemize}

The modifications to the cross-over and mutation operators are presented in the next section, with the assumption that certain terms occur more frequently in a given data origin area. By adopting this approach, we can include ``preferred'' directions for the search space.

\section{Problem statement}

In this section, we present the classical evolutionary approach proposed in \cite{maslyaev2021partial} and its modifications to determine the search space's directions. The approach is essentially a genetic algorithm on a graph, but with an additional local search step using a LASSO operator and subsequent term filtration to compute the fitness function. First, we discuss the key steps of the classical algorithm to define the genotype in Sect.~\ref{sec:classics}a) and the operators in Sect.~\ref{sec:classics}b). Later, we elaborate on the operators' modifications in Sect.~\ref{sec:modified_operators}. Sect.~\ref{sec:scheme} outlines the modifications done to the classical algorithm.

\subsection{Classical evolutionary algorithm}
\label{sec:classics}

\paragraph{Model definition} We operate with building blocks -- tokens that are parametrized families of functions and operators. Token generally has the form shown in Eq.~\ref{eq:token}.

\begin{equation}
    t=t(\pi_1,...\pi_n)
    \label{eq:token}
\end{equation}

In order to distinguish between a single token and a token product (term), we use the notation $T=t_1 \cdot...\cdot t_{T_{length}}$, where $0 < T_{length} \le T_{max}$, and $T_{max}$ is considered an algorithm hyperparameter. However, it is important to note that while $T_{max}$ does affect the final form of the model, a reasonable value (usually 2 or 3) is sufficient to capture most of the actual differential equations. 

Tokens $t_i$ are grouped into token families $\Phi_j$ to aid in fine-tuning the model form. For example, we could define the differential operators family $\Phi_{der}=\{\frac{\partial^{\pi_{n+1}} u}{\partial^{\\pi_1}x_1 ... \partial^{\pi_n}x_n}\}$ to find linear or nonlinear equations with constant coefficients. We could also consider the trigonometric token family $\Phi_{trig}=\{\sin{(\pi_1 x_1+...+\pi_n x_n)}, \cos{(\pi_1 x_1 + ...+\pi_n x_n)}\}$ to search for forcing functions or variable coefficients. The algorithm takes as input the set $\Phi=\bigcup \limits_j \Phi_j$ of chosen or user-defined token families.

We assume that the tokens are pre-computed on a discrete grid for simplicity, but the grid choice does not impact the algorithm's description, so we omit it. It's important to choose the token families for optimization beforehand, as the product of tokens $T$ may contain tokens from different families.

In classical algorithm for term generation, crossover and mutation operators, we introduce set of rules to avoid the trivial cases that are reduced to $0 \equiv 0$ or to avoid the creation of therms that could be obtained with token order change. For clarity, the article only focuses on a differential operator family. 

The differential equation model used during the evolutionary optimization process has the form Eq.~\ref{eq:model}. We note that every equation may be written in this form as soon as term length is sufficient.

\begin{equation}
    M(C_j, \Pi)=\sum \limits_{j=1}^{j \le N_{terms}} C_j T_j
    \label{eq:model}
\end{equation}

In Eq.~\ref{eq:model} with $C_j$ we denote real-valued coefficients and $\Pi=\{\pi_1,...\}$ is the finite set of parameters without fixed length. Every model could have different parameters, and the evolutionary operators could also change this number. A maximal number of terms $N_{terms}$ is the algorithm hyperparameter. We note that the hyperparameter $N_{terms}$ also has not directive but restrictive function. The number of terms in the resulting model may be lower than $N_{terms}$ and is eventually reduced with the fitness calculation procedure described below.

\begin{figure}[h!]
    \centering
    \includegraphics[width=1\linewidth]{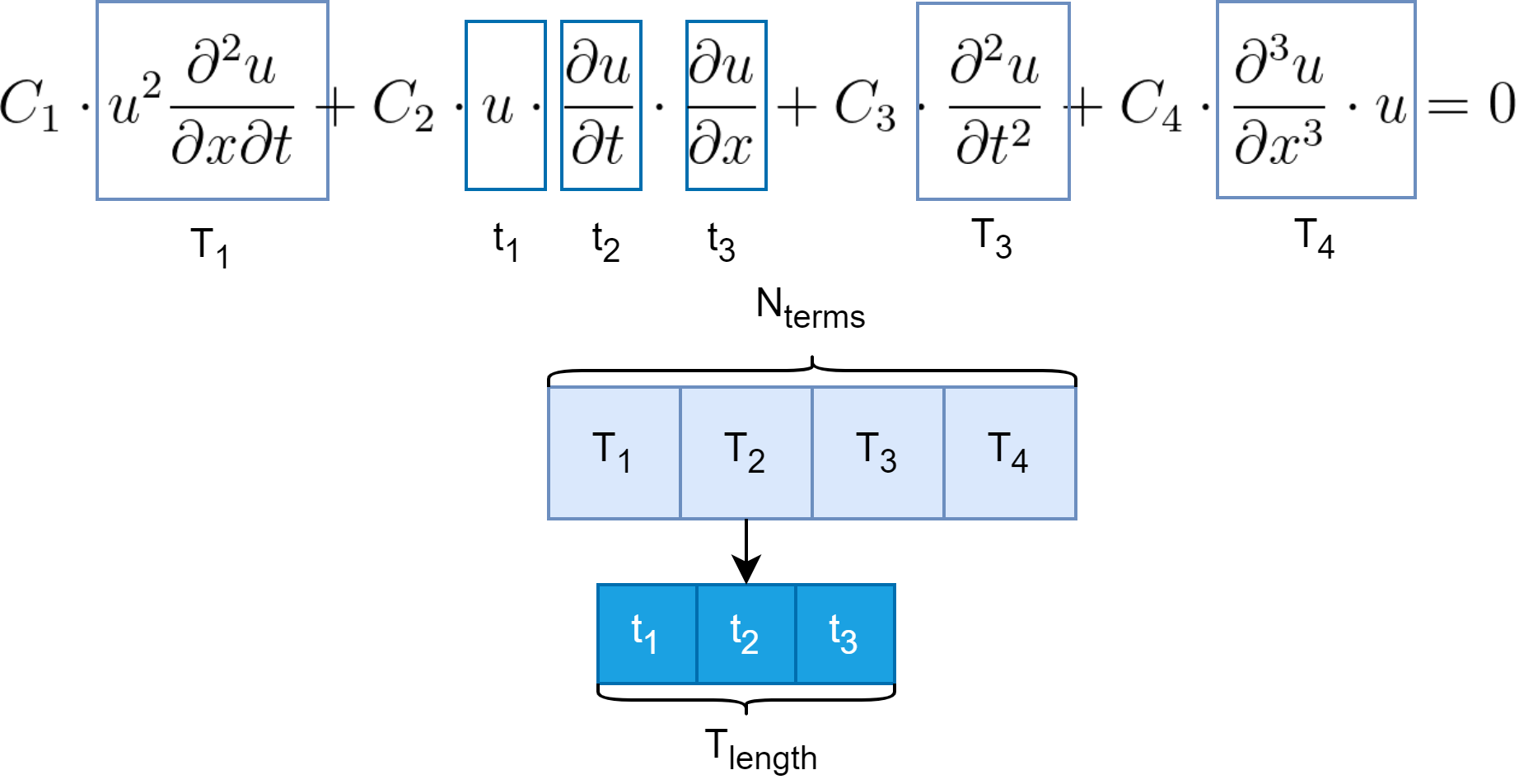}
    \caption{Model visualization: $T_i$ are the token products from Eq.~\ref{eq:model} and $t_i$ are the tokens from Eq.~\ref{eq:token}.}
    \label{fig:model_scheme}
\end{figure}

We use the simplified individual to visualize the following schemes of evolutionary operators, as illustrated in Fig.~\ref{fig:model_scheme}. Each individual corresponds to an instance of the model shown in Eq.~\ref{eq:model}.

\paragraph{Fitness evaluation}

Fitness evaluation has two purposes. First, it allows determining the coefficients $C_j$ for every model -- individ. Second, it serves as a standard measure of individual fitness. One term is randomly chosen as ``target'' for a given model with the parameters set $\Pi$ fixed to evaluate fitness. After that, the LASSO regression is applied to balance the ``target'' term with other terms in the model, as shown in Eq.~\ref{eq:fitness_optimization}.

\begin{multline}
        C_{opt}=\text{arg} \min \limits_{C} \Big|\Big| T_{target}-\sum \limits_{j=1,...T_{target}-1}^{j=T_{target}+1,...,N_{terms}} C_j T_j \Big|\Big|_2 +\\
        + \lambda || C ||_1  , \; C=\{ C_1,...,C_{N_{terms}} \}
        \label{eq:fitness_optimization}
\end{multline}

In Eq.~\ref{eq:fitness_optimization}, with $||\cdot||_p$ corresponding $l_p$ norm is designated. In addition, we note that $C^{(opt)}_{target}=-1$. After applying the LASSO regression operator, the coefficients are compared with the minimal coefficient value threshold of the term. If the absolute value of coefficient $C_j$ is lower than the threshold, then the term is removed from the current model. Thus, the model is refined to reduce the excessive growth of unnecessary terms.

Randomly chosen target allows to avoid trivial solution $\forall j \; C_j=0$. Due to extended variability, we cannot use fixed form as is done in the gradient methods.

After the final set of optimal coefficients for Eq.~\ref{eq:fitness_optimization} is found, the fitness function $F$ is computed as shown in Eq.~\ref{eq:fitness_computation}.

\begin{equation}
F= \frac{1}{||M(\Pi,C_{opt})||_2}  
    \label{eq:fitness_computation}
\end{equation}

\paragraph{Evolutionary operators}

Population initialization and cross-over and mutation operators use a set of rules for generation and exchange. As stated above, the rules are used to avoid situations $0=0$ (for example, two terms that are obtained using the commutative multiplication property are restricted to appear) or to appear of two equal terms during the mutation and cross-over steps.

Apart from the rules restrictions, all tokens in the classical algorithm may appear equiprobably during the mutation, and every term may be equiprobably exchanged during the cross-over.

\begin{figure}[h!]
    \centering
    \includegraphics[width=0.9\linewidth]{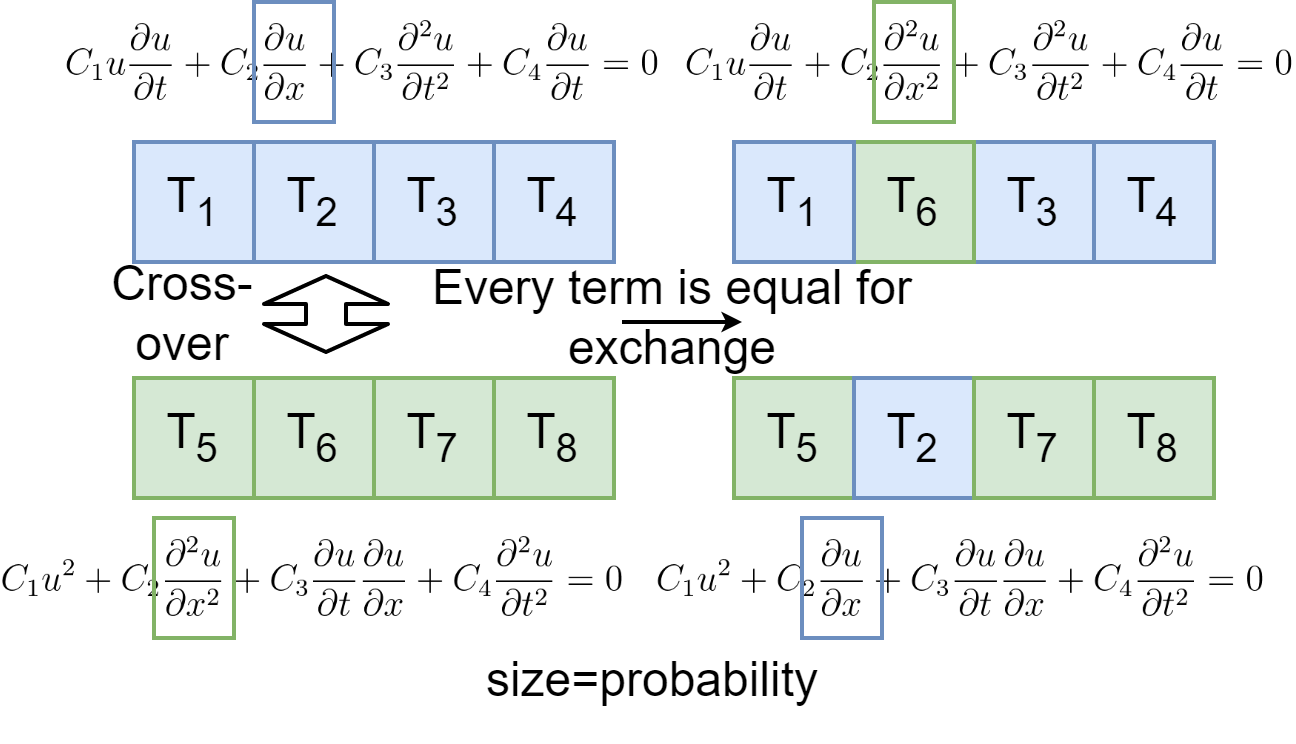}
    \caption{The classical algorithm cross-over. All terms have the same probability of participating in the cross-over.}
    \label{fig:uniform_cross_over}
\end{figure}

The cross-over operator is defined as an exchange of terms between individuals, as shown in Fig.~\ref{fig:uniform_cross_over}. We note that the terms for exchange are chosen using a uniform distribution, i.e. all terms have the same possibility to participate in the exchange.

\begin{figure}[h!]
    \centering
    \includegraphics[width=0.9\linewidth]{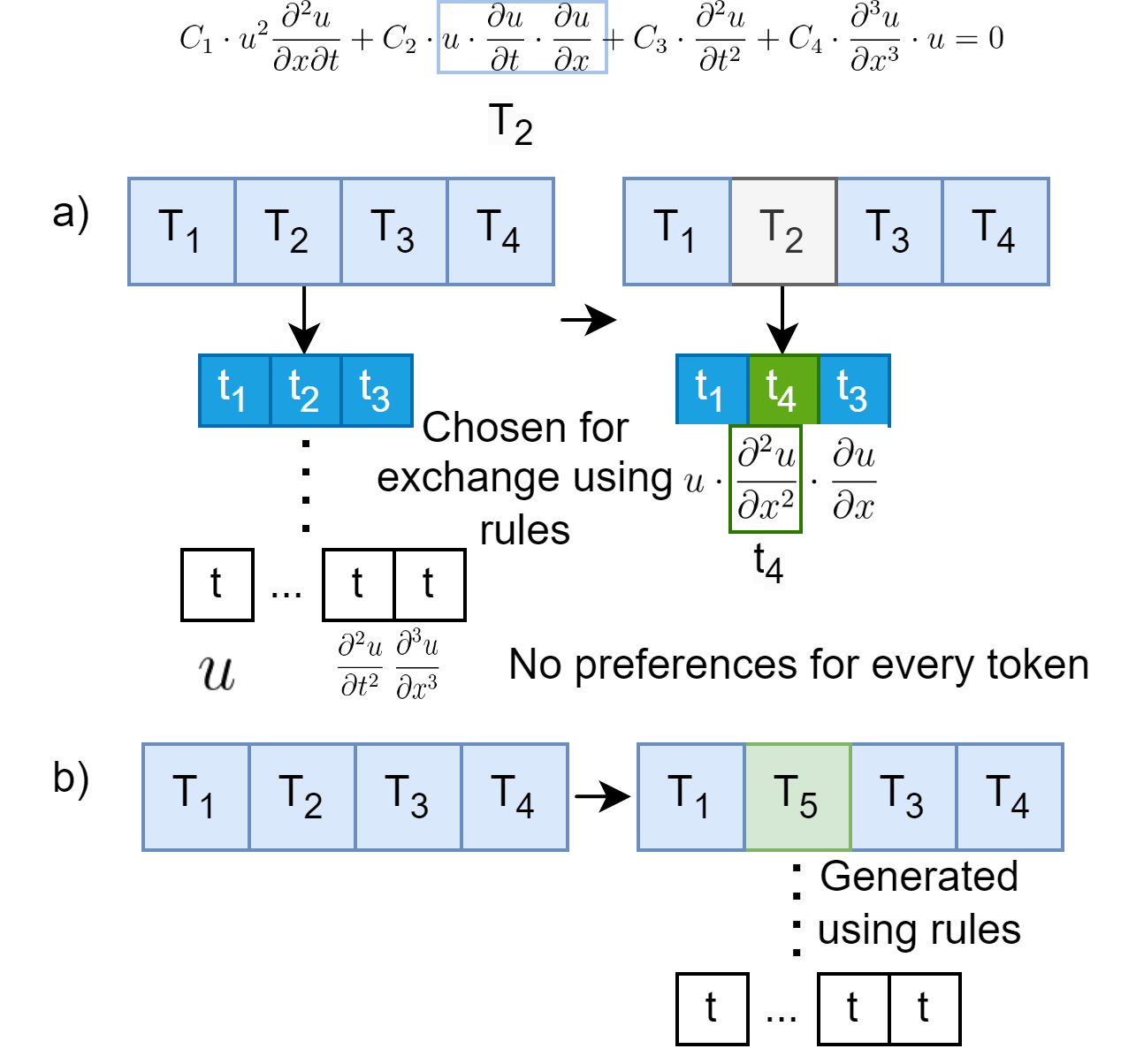}
    \caption{The classical algorithm mutation. New tokens a) and new term b) are generated using a uniform distribution.}
    \label{fig:uniform_mutation}
\end{figure}

The mutation operator has two forms -- term exchange and token exchange -- that could be applied with a given pre-defined probability, as shown in Fig.~\ref{fig:uniform_mutation}.

Token exchange shown in Fig.~\ref{fig:uniform_mutation}a) is simply replacing one token with another using the homogeneous pool of the tokens. For the term exchange (Fig.~\ref{fig:uniform_mutation}b), the new term is generated using the homogeneous pool of tokens: first, the length of the token is chosen randomly and second, the tokens are chosen from the pool. 

The classical algorithm is realized as an evolutionary optimization framework \cite{maslyaev2021multi,maslyaev2022solver}, and the following modifications are also performed as part of a framework modification.

\subsection{Modified evolutionary operators}
\label{sec:modified_operators}

To assume the direction of the search, we add the ``importance'' value to each token. Instead of rules algorithm uses the probability space to pick the tokens for the mutation or the largest probability of the token in the term for the cross-over. During the initial population generation or after the mutation and cross-over operators are applied, the equation is assessed for correctness using a simplified checklist.

It is assumed that the ``importance'' distribution is known. However, we note that extraction of this distribution from only data is an autonomous research problem.

\begin{figure}[h!]
    \centering
    \includegraphics[width=0.85\linewidth]{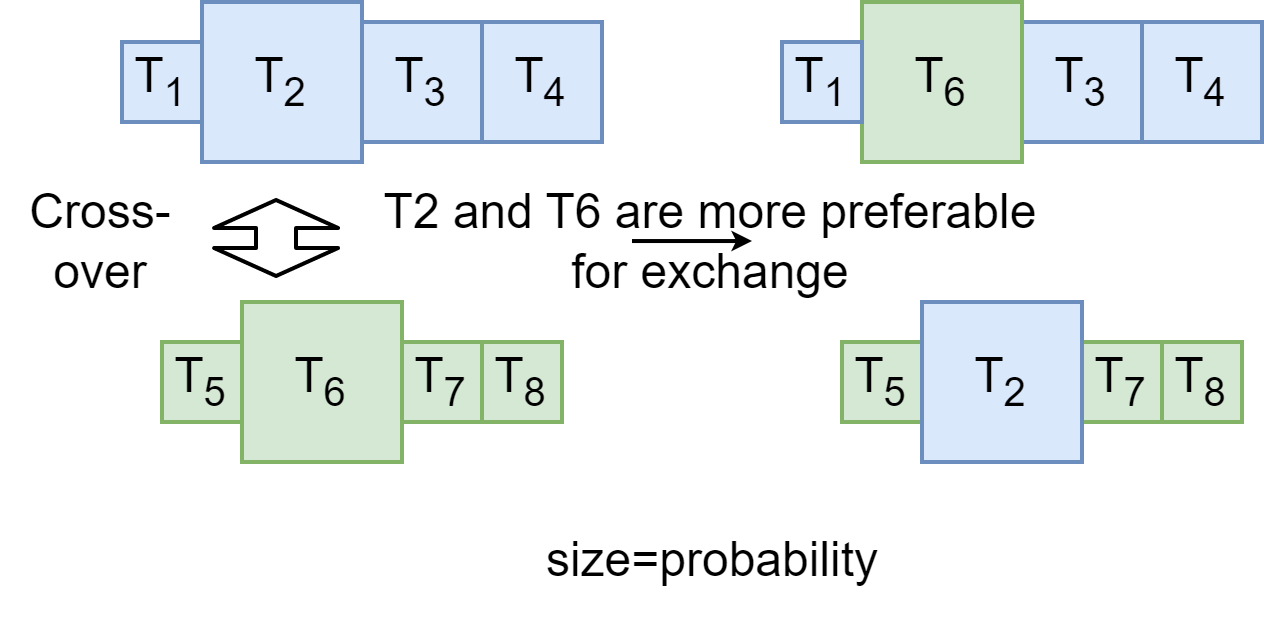}
    \caption{Modified cross-over. Terms have a different probability of participating in the cross-over; for illustration, the most probable terms win.}
    \label{fig:modified_cross_over}
\end{figure}

To better control the optimization process, we use token probability space only in specific locations of the mutation and cross-over operators. The modified cross-over operators take largest token importance as a measure of importance for a given term. Terms with higher importance have a higher chance of participating in cross-over exchange, as shown in Fig.~\ref{fig:modified_cross_over}.

\begin{figure}[h!]
    \centering
    \includegraphics[width=0.85\linewidth]{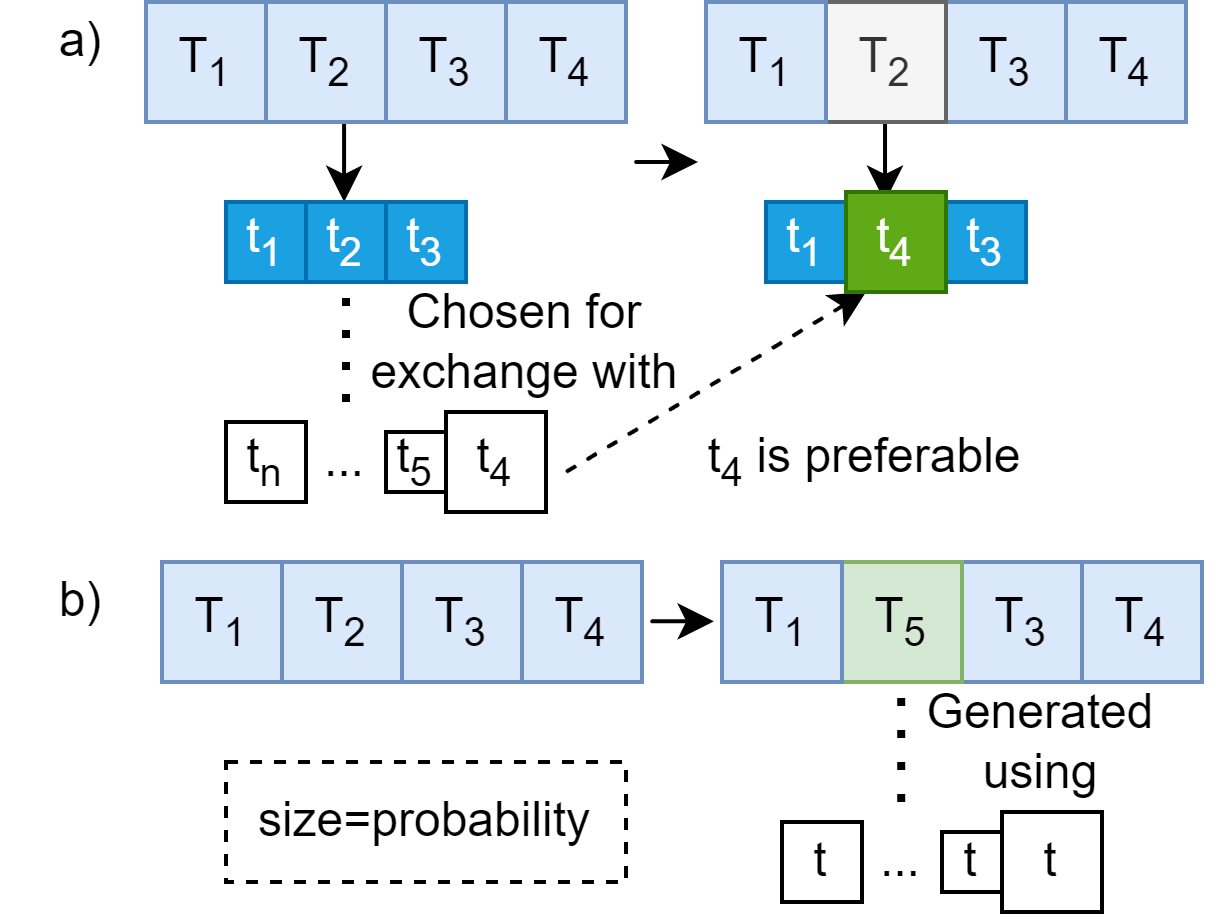}
    \caption{Modified mutation. New token a) is chosen using the importance distribution (for illustration, the most probable token is taken) and new term b) is generated using the importance distribution.}
    \label{fig:modified_mutation}
\end{figure}

The mutation operator takes all terms in the model uniformly, but for a token mutation (Fig.\ref{fig:modified_mutation} a)), its importance is taken into account. Furthermore, for term mutation, a new term is generated using importance distribution as shown in Fig.~\ref{fig:modified_mutation} b).

The term importance is considered as input data. In detail, we allow increasing importance for only a few terms. In this case, all other terms are considered equiprobable, and their importance is computed using the total probability property. The total probability must equal one if all possible terms' importance is summed up.

\subsection{General algorithm scheme}
\label{sec:scheme}

\begin{figure}[h!]
    \centering
    \includegraphics[width=0.8\linewidth]{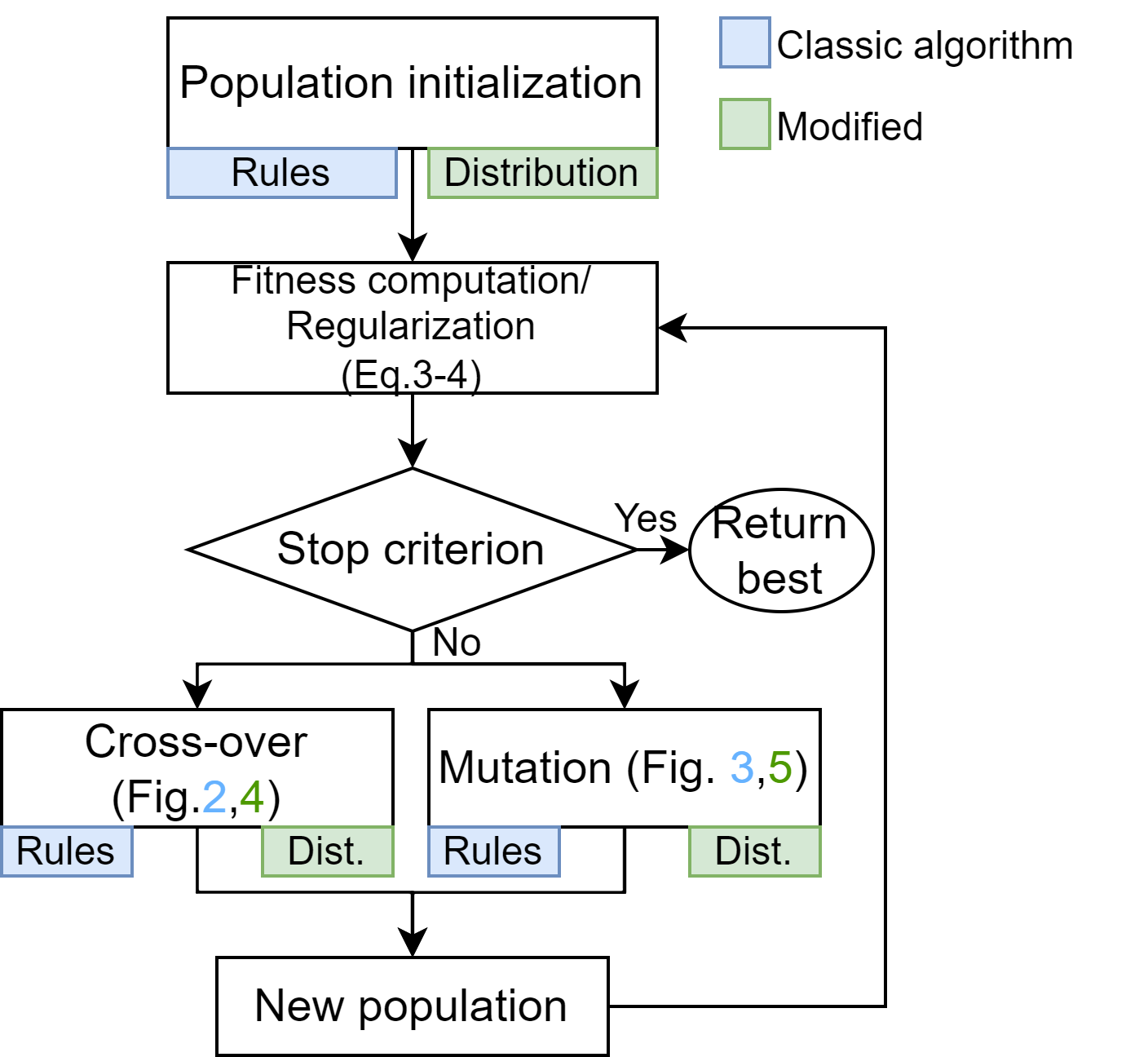}
    \caption{Equation discovery general scheme}
    \label{fig:algorithm_scheme}
\end{figure}

To sum up, the classical and modified algorithms have the same structure as shown in Fig.~\ref{fig:algorithm_scheme}.

We note once more that the main changes are that the rules are replaced with token generator that uses distribution given as input as the rules replacement. The way to obtain this distribution and how it affects the resulting equation is described in the next section.

\section{Experimental results}
\label{sec:experiments}

In this section, the results of several experiments are provided. The influence of equation terms' initial distribution on the evolutionary algorithm performance is examined: first, the classical algorithm is considered; then, mutation operator modification is applied; consequently, terms have a particular fixed initial distribution, from which they are generated when mutation operation occurs; next, the value of importance for desired terms is significantly increased in comparison with the second case; after that, the probability of these terms is highly raised; lastly, the impact of uniformly distributed terms on the performance of the algorithm is explored. The experiments are conducted with inviscid Burgers', wave, and Korteweg -- de Vries equations.

\subsection{Experimental setup}
\label{sec:experimental_setup}

A fixed initial distribution to mimic the classical algorithm rules set is obtained using a term-generating policy described in Alg.~\ref{alg:term_generate}. The generating policy is used to determine terms frequencies, which are in turn used to create a fixed initial distribution.

\begin{algorithm}[h!]
\label{alg:term_generate}
\KwData{maximum number of factors in a term $max\_factors\_in\_term$}
\KwResult{a unique term}
Randomly generate the number of factors in a term: $factors\_num = Uniform (\{1, …, max\_factors\_in\_term\})$ \;
\Do{the term is not unique in an equation}{
    \For{$factor\_idx = 1$ to $factors\_num$}{
    Check structural restrictions (for instance, whether a factor already exists in a term)\;
    Choose a family of functions (e.g. derivatives, trigonometric, polynomial), from which a factor will be generated. The probability of selecting a certain family is proportional to the number of factors that belong to the family\; 
    Randomly choose a factor that belongs to selected family $F$: $factor = Uniform(\{f, f \in F\})$\;
    Update the structural restrictions.
    }
Check the uniqueness of the term\;
}
 \caption{The pseudo-code of term generation}
\end{algorithm}

The probabilities that are obtained using Alg.~\ref{alg:term_generate} are refered to as ``fixed initial distribution'' (i). We note again that for a classical version of the algorithm, the distribution of terms' importance is not uniform due to the introduced equation-building rules that allow for avoiding trivial combinations. The distribution was determined using ten subsequent runs of the classical version of the algorithm.

We calculate the probability of token appearance by dividing the number of records where it occurs (count) by the total number of records. To increase the likelihood of a given token appearing, we adjust the count and re-calculate the probabilities accordingly.

For each equation, we increase the count of terms present in the equation by 20\%, referred to as the ``biased initial distribution'' (ii), and twice this count, referred to as the ``highly biased (initial) distribution'' (iii). The uniform distribution (iv) is achieved by equalizing the appearance of every possible term, resulting in the inclusion of many non-existent equations. However, this serves as a baseline algorithm.

For every experiment, we run ten iterations of the classical algorithm and each of the four types of distribution. When possible, we use pySINDy on the same data. Since this algorithm is gradient-based, results do not vary between runs, but we ensure ten runs to confirm convergence.

Each equation obtained is algebraically transformed to the form of the producing equation plus the noise terms. Runs where transformation to such form is impossible are labeled 'n/a'. To compare our results, we calculate the mean absolute error (MAE) between the coefficients of the 'ground truth' equation and the resulting model, without noise terms. This comparison metric is used in the equation discovery area \cite{fasel2022ensemble}. Additionally, we consider convergence time as a secondary comparison metric.

With the experiments, we check that the distribution that mimics the rules of the classical algorithm is a good example of an extracted knowledge. If we increase the importance of the given equation terms, we naturally increase the resulting equation quality, however, up to some extent. That is, case (ii) will show the best possible result overall.

\subsection{Burgers' equation}

The initial-boundary value problem for Burger's equation is shown in Eq.~\ref{eq:burgers_eq}.

\begin{equation}
\begin{array}{cc}
     \frac{\partial u}{\partial t} + u \frac{\partial u}{\partial x} = 0  \\
     u(0,t)=\begin{cases}
    1000, t\geq 2 \\
    0, t < 2
\end{cases}\\
u(x,0)=\begin{cases}
    1000, x \leq - 2000 \\
    -x/2, -2000<x<0 \\
    0, \text{otherwise}
\end{cases}\\
(x,t) \in [-4000,4000] \times [0,4]
\end{array}
    \label{eq:burgers_eq}
\end{equation}

The boundary problem Eq.~\ref{eq:burgers_eq} has an analytical solution \cite{rudy2017data}. Data were obtained using an analytical solution and discretization of a computation domain $(x,t) \in [-4000,4000] \times [0,4]$ using $101 \times 101$ points. The derivatives are also taken using an analytical solution and discretized using the same grid to reduce the influence of the numerical differentiation algorithm.

\begin{figure}[h!]
    \centering
    \includegraphics[width=0.9\linewidth]{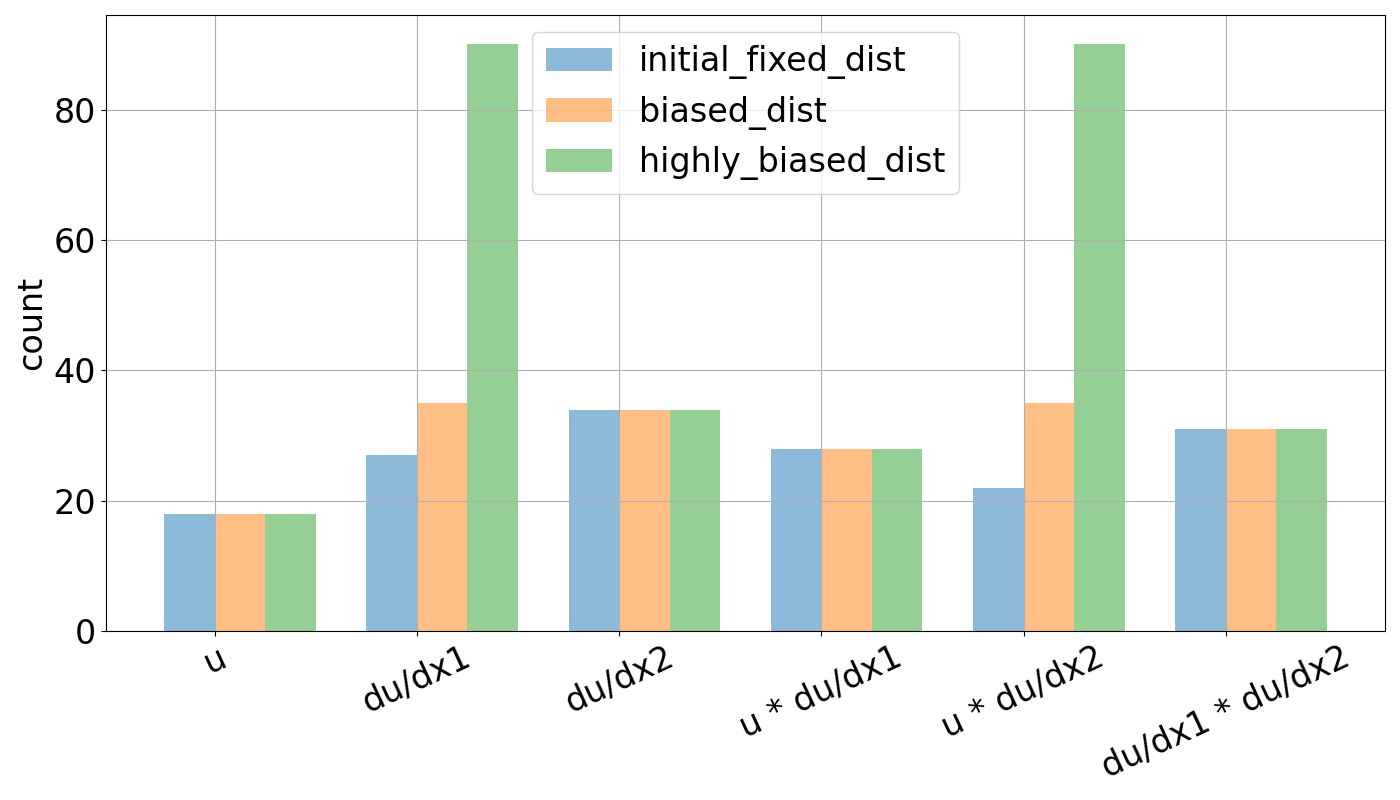}
    \caption{Importance distributions forms for cases classical algorithm (i, blue), the classical algorithm with a moderate increase (ii, orange) and significant increase (iii, green) of the Burgers' equation terms importance.}
    \label{fig:distsburgers}
\end{figure}

For experiments, we run classical algorithm and modified with four generated distributions as described in Sec.~\ref{sec:classics}. Selected terms' counts for Burgers' equation are shown in Fig.~\ref{fig:distsburgers}.

The experimental results for ten subsequent runs for each distribution are shown in Tab.~\ref{tab:burgers_mae}. We note that we could not restore the correct equation with the pySINDy algorithm using the data obtained with Eq.~\ref{eq:burgers_eq}.

\begin{table}[h!]
\centering
\caption{Coefficients MAE, Burgers' equation. The green color is used to mark the minimal error obtained.}
\label{tab:burgers_mae}
\begin{tabular}{|c|c|c|c|c|}
\hline
\begin{tabular}[c]{@{}c@{}}Basic \\ algorithm\end{tabular} & \begin{tabular}[c]{@{}c@{}}Fixed \\ initial \\ distr.\end{tabular} & \begin{tabular}[c]{@{}c@{}}Biased \\ initial \\ distr.\end{tabular} & \begin{tabular}[c]{@{}c@{}}Highly\\ biased\\ distr.\end{tabular} & \begin{tabular}[c]{@{}c@{}}Uniform\\ initial\\ distr.\end{tabular} \\ \hline
0.0134                                                     & 0.0134                                                             & 0.0134                                                              & \cellcolor[HTML]{C5E0B3}0.0009                                   & 0.0134                                                             \\ \hline
0.0134                                                     & \cellcolor[HTML]{C5E0B3}0.001                                      & 0.0134                                                              & 0.0134                                                           & \cellcolor[HTML]{C5E0B3}0.001                                      \\ \hline
\cellcolor[HTML]{C5E0B3}0.0009                             & \cellcolor[HTML]{C5E0B3}0.001                                      & \cellcolor[HTML]{C5E0B3}0.0009                                      & \cellcolor[HTML]{C5E0B3}0.001                                    & 0.0134                                                             \\ \hline
0.0134                                                     & 0.0134                                                             & \cellcolor[HTML]{C5E0B3}0.0009                                      & \cellcolor[HTML]{C5E0B3}0.001                                    & \cellcolor[HTML]{C5E0B3}0.0009                                     \\ \hline
0.0134                                                     & 0.0134                                                             & 0.0134                                                              & \cellcolor[HTML]{C5E0B3}0.0009                                   & 0.0134                                                             \\ \hline
0.0134                                                     & 0.0134                                                             & \cellcolor[HTML]{C5E0B3}0.0009                                      & 0.0134                                                           & \cellcolor[HTML]{C5E0B3}0.001                                      \\ \hline
\cellcolor[HTML]{C5E0B3}0.001                              & \cellcolor[HTML]{C5E0B3}0.001                                      & \cellcolor[HTML]{C5E0B3}0.001                                       & \cellcolor[HTML]{C5E0B3}0.001                                    & 0.0134                                                             \\ \hline
0.0134                                                     & 0.0134                                                             & \cellcolor[HTML]{C5E0B3}0.001                                       & \cellcolor[HTML]{C5E0B3}0.001                                    & 0.0134                                                             \\ \hline
0.0134                                                     & 0.0134                                                             & \cellcolor[HTML]{C5E0B3}0.0009                                      & \cellcolor[HTML]{C5E0B3}0.001                                    & \cellcolor[HTML]{C5E0B3}0.001                                      \\ \hline
0.0134                                                     & 0.0134                                                             & 0.0134                                                              & 0.0134                                                           & 0.0134                                                             \\ \hline
\end{tabular}
\end{table}

We note that as a result only several types of ``significant part'' of the equation with different noise terms are obtained. After noise is truncated the equations usually may be grouped into two of three MAE clusters. As a result, for the Burgers equation case distribution (ii) ``biased fixed distribution'', provides the best result in terms of MAE. 

\begin{figure}[h!]
    \centering
    \includegraphics[width=0.9\linewidth]{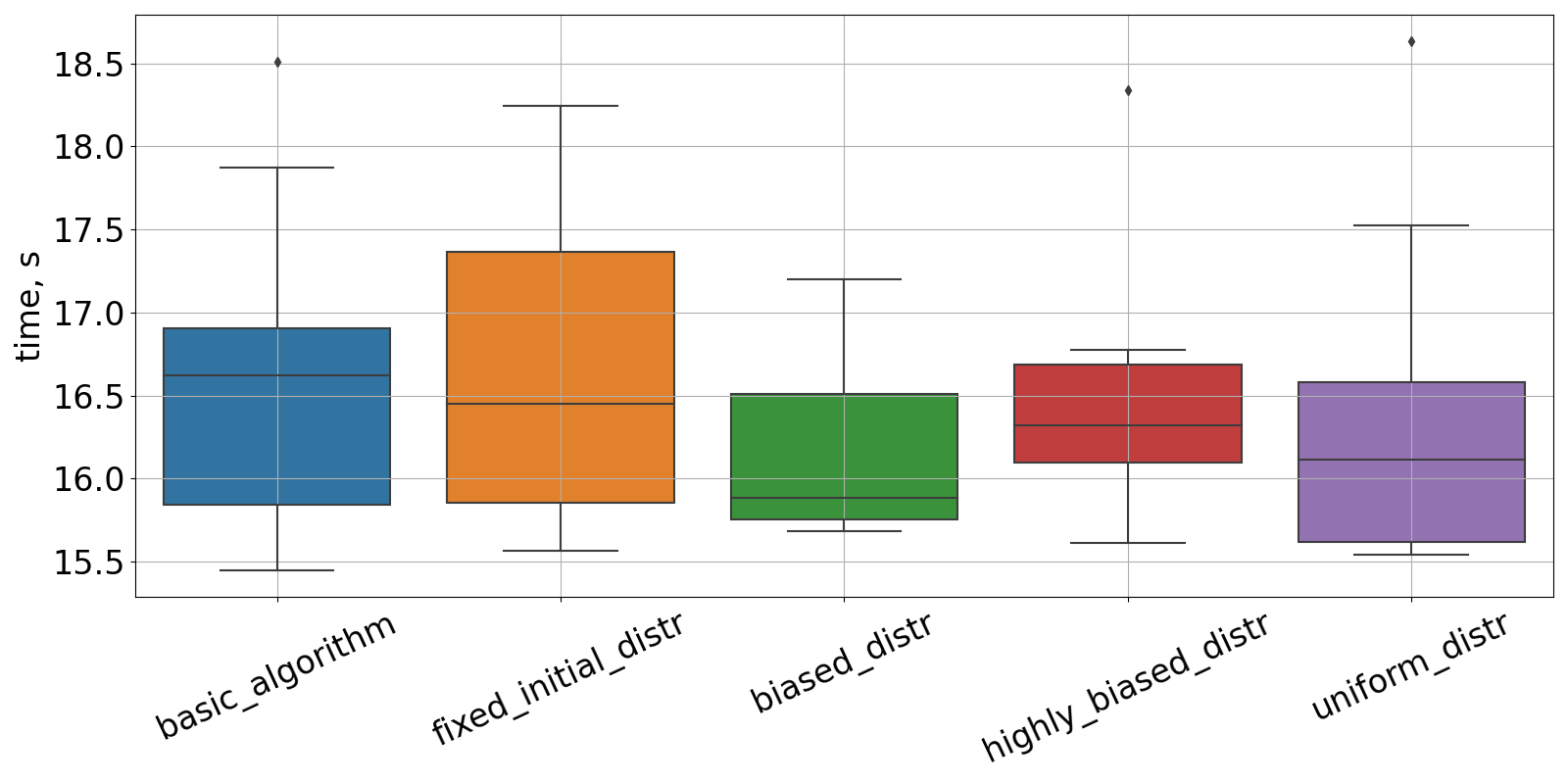}
    \caption{Algorithm running time, Burgers' equation from left to right: classical algorithm, modified algorithm with classical term importance distribution, moderate and significant Burger's term importance increase, uniform distribution.}
    \label{fig:time_burgers}
\end{figure}

A decrease in variability introduced by the biased distribution may negatively affect the algorithm's convergence time. For each run, the optimization time was recorded and shown in Fig.~\ref{fig:time_burgers}.  

As seen in Fig~\ref{fig:time_burgers}, the optimization time is affected insignificantly. Therefore, we conclude that the moderately biased importance distribution (ii, orange, Fig.~\ref{fig:distsburgers}) is the best possible distribution for Burgers' equation.

\subsection{Wave equation}

Wave equation only occasionally appears in equation discovery applications since the equation form is usually restricted to the $u_t=F(x,u_x,u_xx,...)$. However, the described evolutionary approach allows for wave equation restoration. The initial-boundary value problem for the wave equation is shown in Eq.~\ref{eq:wave_eq}.

\begin{equation}
\begin{array}{cc}
\frac{\partial^2 u}{\partial t^2} - \frac{1}{25} \frac{\partial^2 u}{\partial x^2} = 0 \\
u(0,t)=u(1,t)=0\\
u(x,0)=10^4 \sin^2{\frac{1}{10}x(x-1)}\\
u'(x,0)=10^3 \sin^2{\frac{1}{10}x(x-1)}\\
(x,t) \in [0,1] \times [0,1]
\end{array}
    \label{eq:wave_eq}
\end{equation}

The solution was obtained numerically using Wolfram Mathematica software. It provides an interpolated solution, which was computed using $101 \times 101$ discretization points in the domain $(x,t) \in [0,1] \times [0,1]$. The derivatives were obtained using interpolated solution differentiation within Wolfram Mathematica software.

For experiments, we run classical algorithm and modified with four generated distributions as described in Sec.~\ref{sec:classics}. Selected terms' counts for wave equation are shown in Fig.~\ref{fig:distsburgers}.

\begin{figure}[h!]
    \centering
    \includegraphics[width=0.9\linewidth]{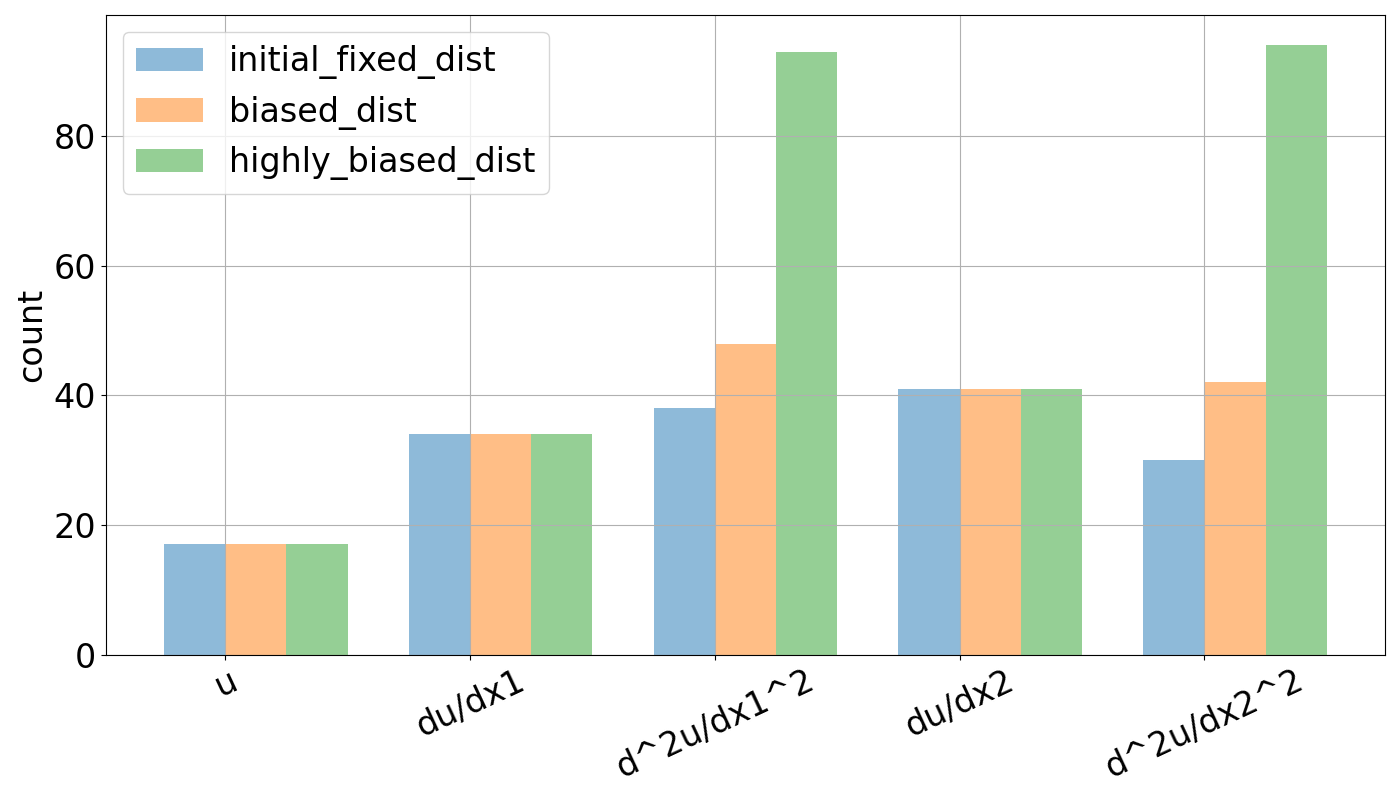}
    \caption{Importance distributions forms for cases classical algorithm (i, blue), the classical algorithm with a moderate increase (ii, orange) and significant increase (iii, green) of the wave equation terms importance.}
    \label{fig:distswave}
\end{figure}

The experimental results are shown in Tab.~\ref{tab:wave_mae}.  In this case, the wave equation structure was not recovered for every run. The runs for which the wave equation was not recovered are shown as 'n/a' in Tab.~\ref{tab:wave_mae}. We note that the pySINDy algorithm has fixed equation form $u_t=F(u,u_x,u_{xx},...)$, and thus the wave equation may not be obtained without the code changes.

\begin{table}[h!]
\centering
\caption{Coefficients MAE, wave equation. The green color marks minimal obtained error, and 'n/a' marks run when the algorithm cannot restore the equation with given importance distribution.}
\label{tab:wave_mae}
\begin{tabular}{|c|c|c|c|c|}
\hline
\begin{tabular}[c]{@{}c@{}}Basic \\ algorithm\end{tabular} & \begin{tabular}[c]{@{}c@{}}Fixed \\ initial\\ distr.\end{tabular} & \begin{tabular}[c]{@{}c@{}}Biased\\ initial\\ distr.\end{tabular} & \begin{tabular}[c]{@{}c@{}}Highly\\ biased\\ distr.\end{tabular} & \begin{tabular}[c]{@{}c@{}}Uniform\\ initial\\ distr.\end{tabular} \\ \hline
\cellcolor[HTML]{C5E0B3}0.0086                             & \cellcolor[HTML]{C5E0B3}0.0086                                    & \cellcolor[HTML]{C5E0B3}0.0086                                    & 0.0423                                                           & \cellcolor[HTML]{C5E0B3}0.009                                      \\ \hline
\cellcolor[HTML]{C0C0C0}n/a                                & \cellcolor[HTML]{C5E0B3}0.0086                                    & \cellcolor[HTML]{C5E0B3}0.0086                                    & 0.0423                                                           & \cellcolor[HTML]{C5E0B3}0.009                                      \\ \hline
\cellcolor[HTML]{C5E0B3}0.0086                             & \cellcolor[HTML]{C0C0C0}n/a                                       & \cellcolor[HTML]{C5E0B3}0.0086                                    & 0.0423                                                           & \cellcolor[HTML]{C5E0B3}0.0086                                     \\ \hline
\cellcolor[HTML]{C5E0B3}0.0086                             & \cellcolor[HTML]{C5E0B3}0.0086                                    & \cellcolor[HTML]{C5E0B3}0.0086                                    & 0.0423                                                           & \cellcolor[HTML]{C5E0B3}0.0086                                     \\ \hline
\cellcolor[HTML]{C5E0B3}0.0086                             & \cellcolor[HTML]{C5E0B3}0.0086                                    & \cellcolor[HTML]{C5E0B3}0.0086                                    & \cellcolor[HTML]{C0C0C0}n/a                                      & 0.0423                                                             \\ \hline
\rowcolor[HTML]{C5E0B3} 
0.0086                                                     & 0.0086                                                            & 0.0086                                                            & 0.0086                                                           & 0.0086                                                             \\ \hline
\cellcolor[HTML]{C5E0B3}0.0086                             & 0.0423                                                            & \cellcolor[HTML]{C5E0B3}0.0086                                    & \cellcolor[HTML]{C5E0B3}0.009                                    & \cellcolor[HTML]{C5E0B3}0.0086                                     \\ \hline
\cellcolor[HTML]{C5E0B3}0.0086                             & \cellcolor[HTML]{C5E0B3}0.0086                                    & \cellcolor[HTML]{C5E0B3}0.0086                                    & 0.0423                                                           & \cellcolor[HTML]{C0C0C0}n/a                                        \\ \hline
\rowcolor[HTML]{C5E0B3} 
0.0086                                                     & 0.0086                                                            & 0.009                                                             & 0.0086                                                           & 0.0086                                                             \\ \hline
\cellcolor[HTML]{C5E0B3}0.0086                             & 0.0423                                                            & \cellcolor[HTML]{C5E0B3}0.0086                                    & \cellcolor[HTML]{C0C0C0}n/a                                      & \cellcolor[HTML]{C5E0B3}0.0086                                     \\ \hline
\end{tabular}
\end{table}

\begin{figure}[h!]
    \centering
    \includegraphics[width=0.9\linewidth]{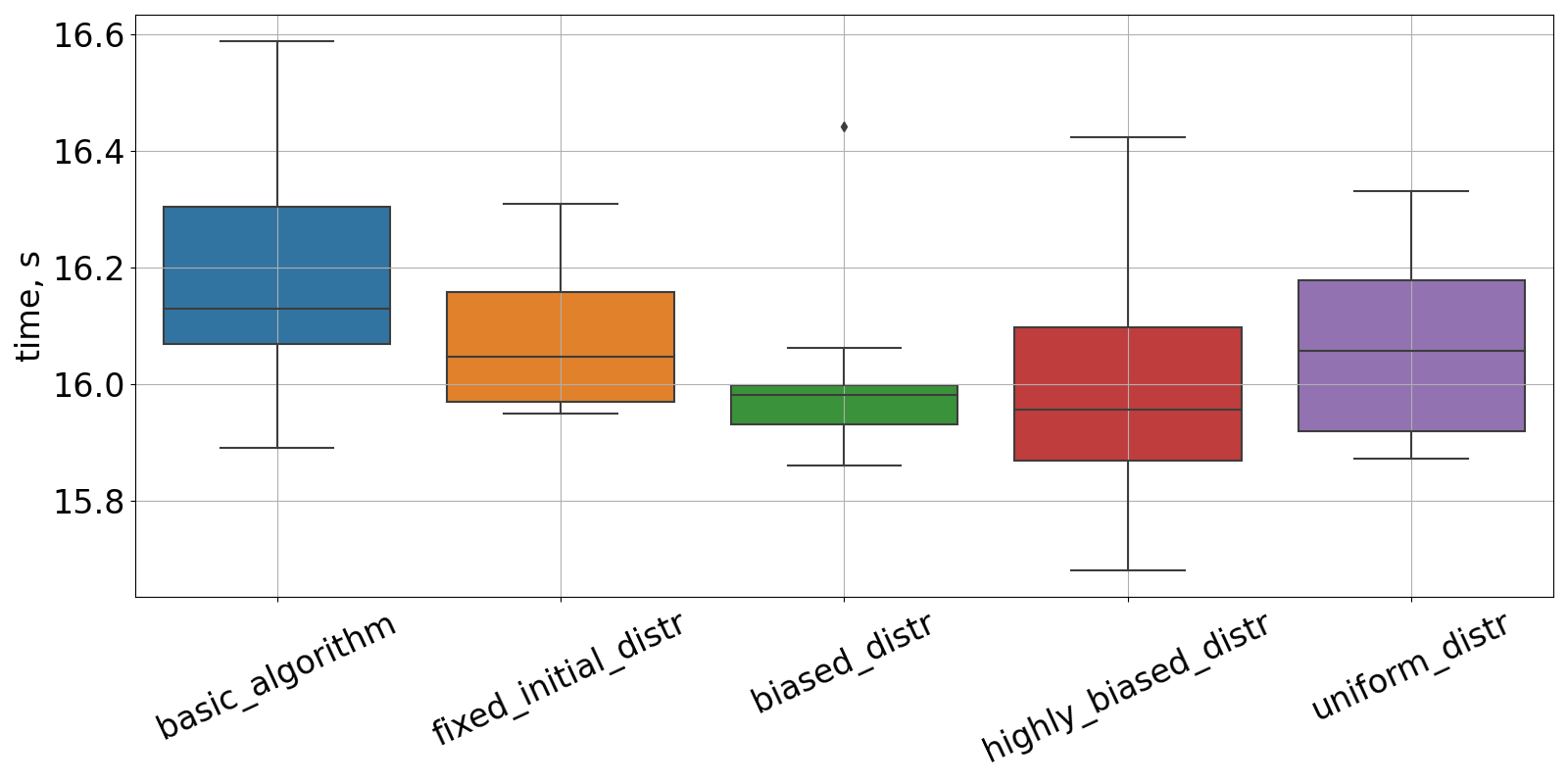}
    \caption{Algorithm running time, wave equation from left to right: classical algorithm, modified algorithm with classical term importance distribution, moderate and significant wave equation terms importance increase, uniform distribution.}
    \label{fig:time_wave}
\end{figure}

The classical algorithm and modified one with distribution (ii) ``biased fixed distribution'' shows almost the same result for the wave equation case. However, the classical algorithm cannot find the correct structure for some runs. The time consumption is also recorded and shown in Fig.~\ref{fig:time_wave}.

We note again that the time spent on optimization varies insignificantly. In contrast, adding the direction to the algorithm for the wave equation case allows finding the correct structure faster and more often than the classical algorithm.

\subsection{Korteweg -- de Vries equation}

Inhomogenous Korteweg-de Vries equation is chosen as a more challenging case. The initial-boundary value problem is shown in Eq.~\ref{eq:KdV}.

\begin{equation}
\begin{array}{cc}
u_t + 6 u u_x + u_{xxx} = \cos t \sin x \\
u(x,0)=0\\
\left[u_{xx}+2 u_x+u \right] \Big|_{x=0}=0\\
\left[2 u_{xx}+u_x+3 u \right] \Big|_{x=1}=0\\
\left[5 u_x+5 u \right] \Big|_{x=1}=0\\
(x,t) \in [0,1] \times [0,1]
\end{array}
    \label{eq:KdV}
\end{equation}

The solution was obtained in the same way as in the case of the wave equation. We use Wolfram Mathematica software and $101 \times 101$ discretization points in the domain $(x,t) \in [0,1] \times [0,1]$. The derivatives were also obtained using interpolated solution differentiation within Wolfram Mathematica software.

\begin{figure}[h!]
    \centering
    \includegraphics[width=0.9\linewidth]{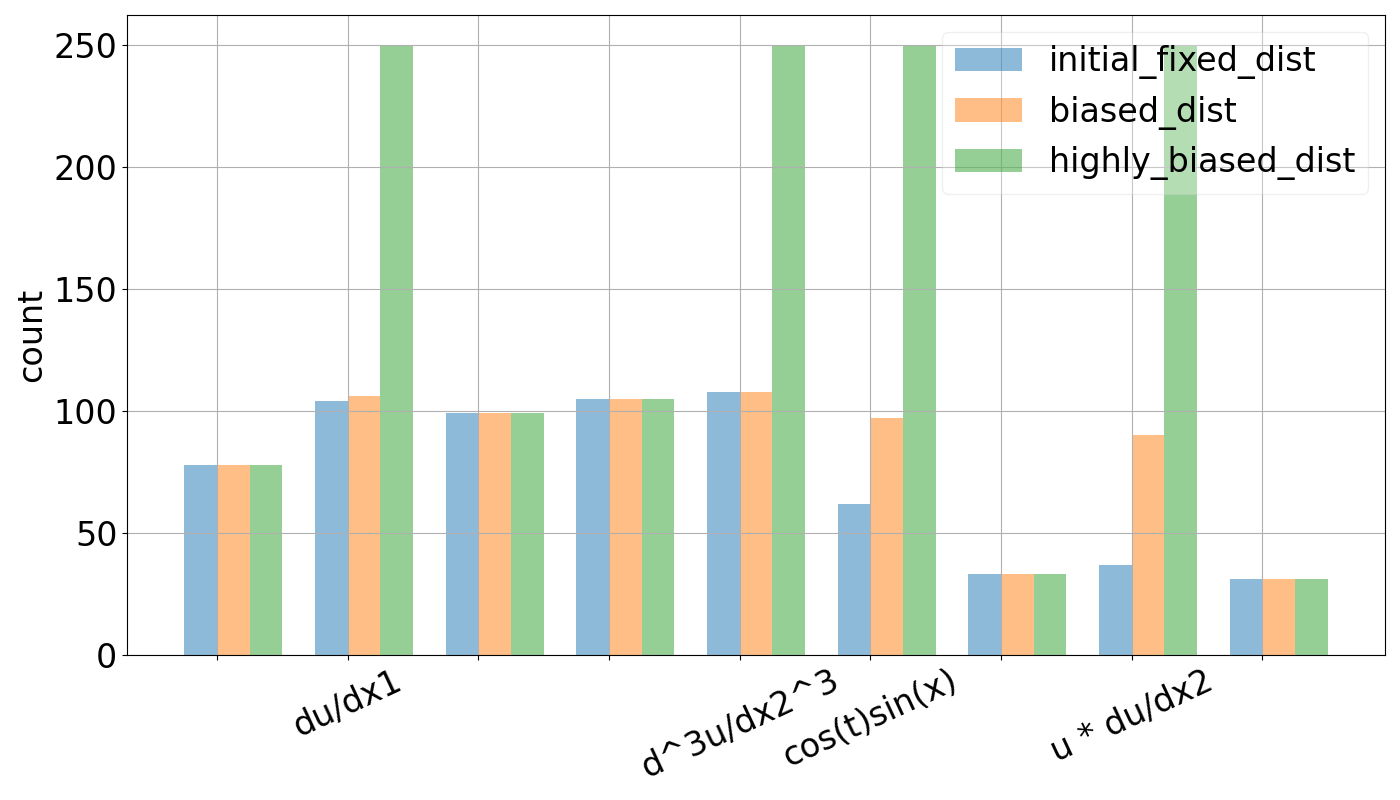}
    \caption{Importance distributions forms for cases classical algorithm (i, blue), the classical algorithm with a moderate increase (ii, orange) and significant increase (iii, green) of the Korteweg -- de Vries equation terms importance.}
    \label{fig:kdv_distribs}
\end{figure}

The experimental setup does not differ from previous cases and contains four possible distributions. However, because of the presence of the third spatial derivative, the overall number of terms was significantly higher. The resulting significance distributions are shown in Fig.~\ref{fig:kdv_distribs}.

Experimental results are shown in Tab.~\ref{tab:kdv_mae}. In this case, the correct structure is found less often and the effect of the distribution appears to be more apparent. The proper distribution choice has the effect of discovering proper structure in more runs. We note that the pySINDy algorithm cannot handle non-homogenous equations by default.

\begin{table}[h!]
\centering
\caption{Coefficients MAE, Korteweg -- de Vries equation. The green color marks the minimal obtained error, and 'n/a' marks run when the algorithm cannot restore the equation with given importance distribution.}
\label{tab:kdv_mae}
\begin{tabular}{|c|c|c|c|c|}
\hline
\begin{tabular}[c]{@{}c@{}}Basic \\ algorithm\end{tabular} &
  \begin{tabular}[c]{@{}c@{}}Fixed\\ initial\\ distr.\end{tabular} &
  \begin{tabular}[c]{@{}c@{}}Biased\\ initial\\ distr.\end{tabular} &
  \begin{tabular}[c]{@{}c@{}}Highly\\ biased\\ distr.\end{tabular} &
  \begin{tabular}[c]{@{}c@{}}Uniform\\ initial\\ distr.\end{tabular} \\ \hline
\rowcolor[HTML]{C0C0C0} 
n/a                            & n/a                         & \cellcolor[HTML]{C5E0B3}0.0001 & \cellcolor[HTML]{C5E0B3}0.0001 & n/a                         \\ \hline
\rowcolor[HTML]{C5E0B3} 
\cellcolor[HTML]{C0C0C0}n/a    & 0.0001                      & \cellcolor[HTML]{C0C0C0}n/a    & 0.0001                         & 0.0001                      \\ \hline
\rowcolor[HTML]{C0C0C0} 
\cellcolor[HTML]{C5E0B3}0.0001 & n/a                         & \cellcolor[HTML]{C5E0B3}0.0001 & n/a                            & n/a                         \\ \hline
\rowcolor[HTML]{C5E0B3} 
0.0001                         & 0.0001                      & \cellcolor[HTML]{C0C0C0}n/a    & 0.0001                         & 0.0001                      \\ \hline
0.0011                         & \cellcolor[HTML]{C0C0C0}n/a & \cellcolor[HTML]{C0C0C0}n/a    & \cellcolor[HTML]{C0C0C0}n/a    & \cellcolor[HTML]{C0C0C0}n/a \\ \hline
\rowcolor[HTML]{C5E0B3} 
0.0001                         & 0.0001                      & 0.0001                         & 0.0001                         & 0.0001                      \\ \hline
\rowcolor[HTML]{C5E0B3} 
0.0001                         & \cellcolor[HTML]{C0C0C0}n/a & 0.0001                         & 0.0001                         & \cellcolor[HTML]{C0C0C0}n/a \\ \hline
\cellcolor[HTML]{C0C0C0}n/a    & \cellcolor[HTML]{C0C0C0}n/a & \cellcolor[HTML]{C5E0B3}0.0001 & 0.0014                         & \cellcolor[HTML]{C0C0C0}n/a \\ \hline
\rowcolor[HTML]{C5E0B3} 
\cellcolor[HTML]{C0C0C0}n/a    & 0.0001                      & 0.0001                         & 0.0001                         & \cellcolor[HTML]{C0C0C0}n/a \\ \hline
\rowcolor[HTML]{C0C0C0} 
n/a                            & n/a                         & \cellcolor[HTML]{C5E0B3}0.0001 & n/a                            & n/a                         \\ \hline
\end{tabular}
\end{table}

\begin{figure}[h!]
    \centering
    \includegraphics[width=0.9\linewidth]{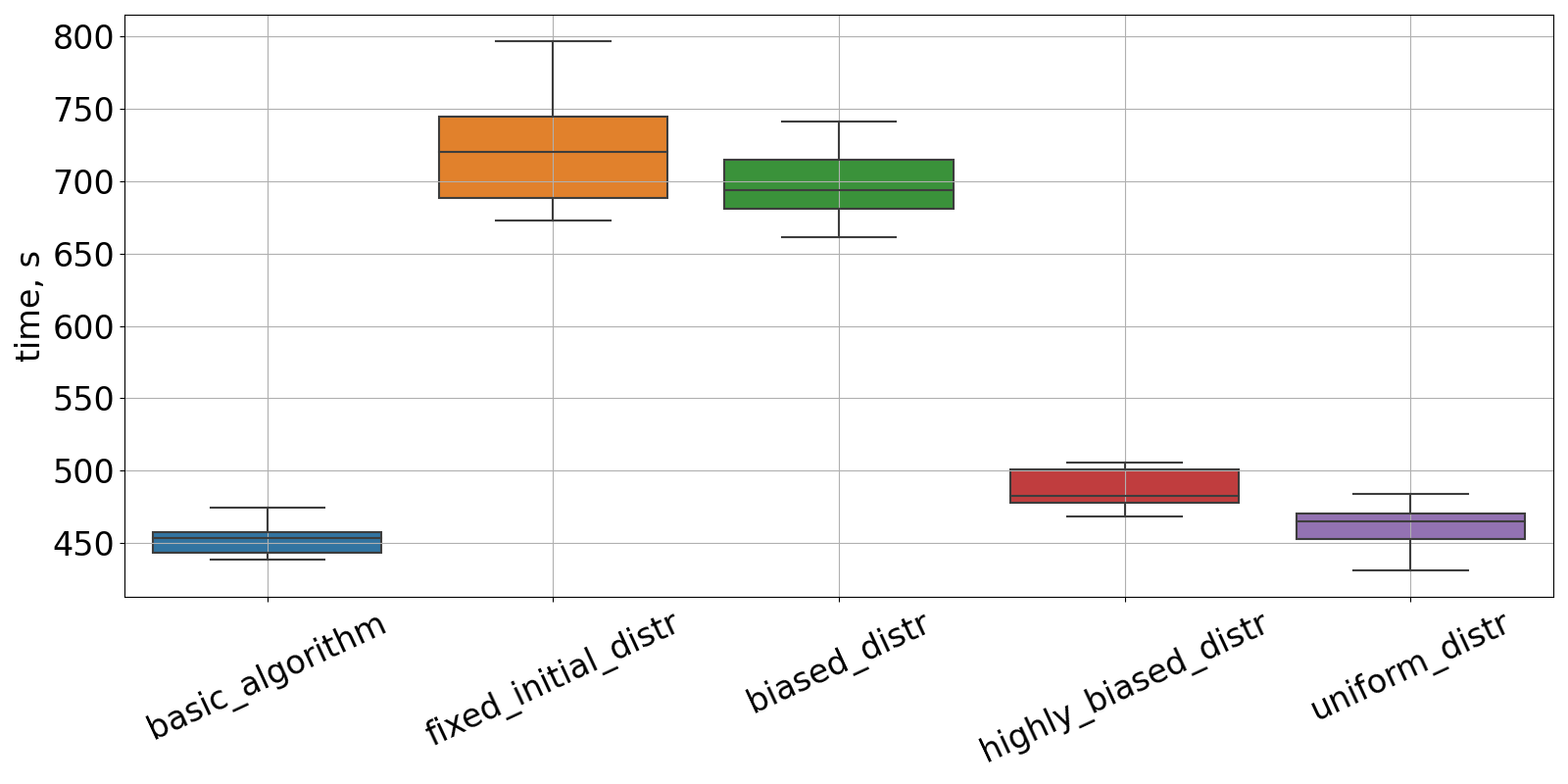}
    \caption{Algorithm running time, Korteweg -- de Vries equation from left to right: classical algorithm, the modified algorithm with classical term importance distribution, moderate and significant Korteweg -- de Vries equation term importance increase, uniform distribution.}
    \label{fig:kdv_time}
\end{figure}

Time assessment (Fig.~\ref{fig:kdv_time}) shows that the time spent to obtain the resulting structure with a distribution distribution (ii) ``biased fixed distribution'' in the Korteweg--de Vries equation case is higher than that used the classical algorithm to converge. However, this distribution allows one to obtain the correct structure more often.

\subsection{Comparison with PySINDy: Burger's equation}

The authors provided the problem and the data \footnote{https://github.com/dynamicslab/pysindy} \cite{desilva2020,Kaptanoglu2022}. The resulting problem can be formulated as in Eq.~\ref{eq:Burgers2}. We note that the boundary conditions are not reported.

\begin{equation}
\begin{array}{cc}
\frac{\partial u}{\partial t} + u \frac{\partial u}{\partial x} - 0.1 \frac{\partial^2 u}{\partial x^2} = 0 \\
(x,t) \in [-8,8] \times [0,10]
\end {array}
    \label{eq:Burgers2}
\end{equation}

\begin{figure}[h!]
    \centering
    \includegraphics[width=0.9\linewidth]{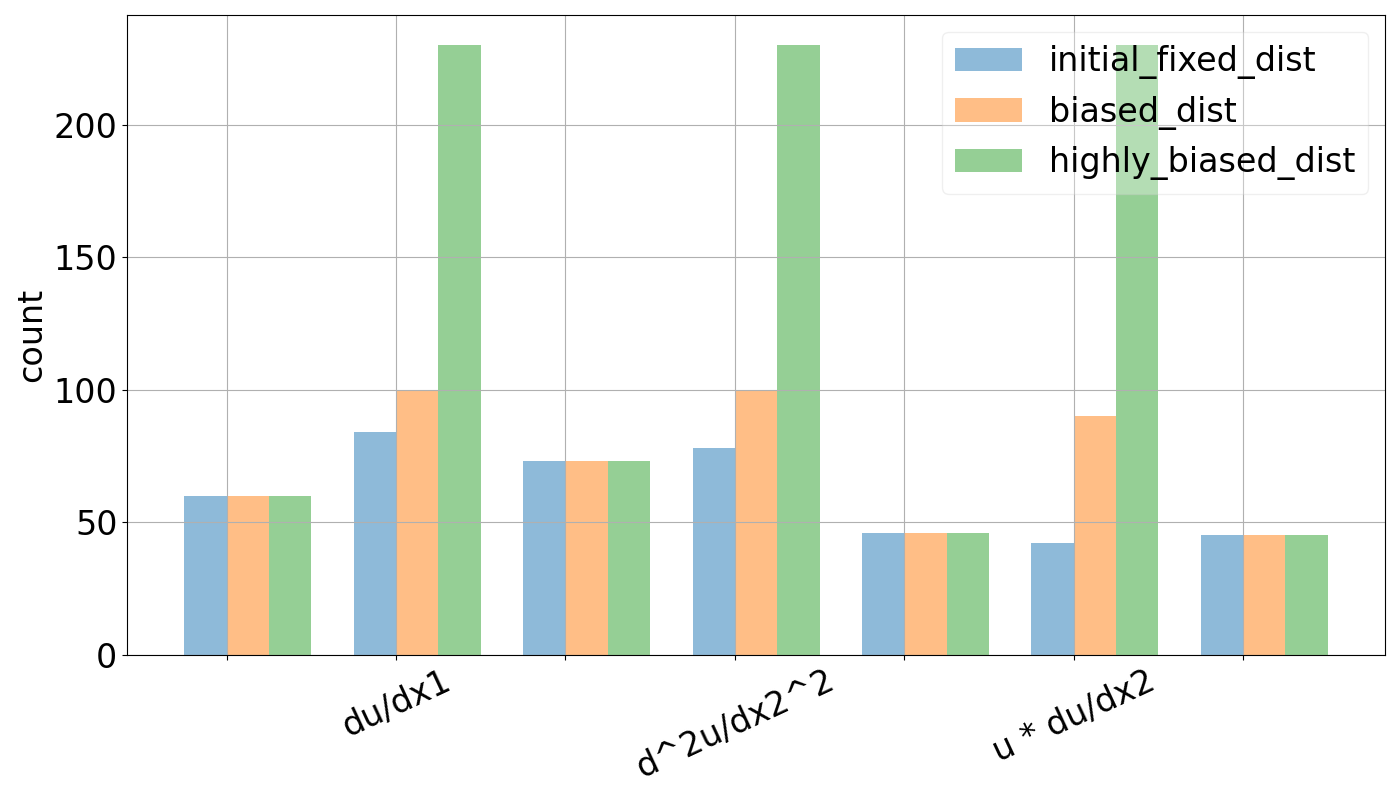}
    \caption{Importance distributions forms for cases classical algorithm (i, blue), the classical algorithm with a moderate increase (ii, orange) and significant increase (iii, green) of the Burger's equation terms importance.}
    \label{fig:distsburg2}
\end{figure}

The authors found the solution on the domain $(x,t) \in [-8,8] \times [0,10]$ using $ 256 \times 101$ discretization points.
The importance distributions for Burger's equation are shown in Fig.~\ref{fig:distsburg2}.

The results of the experiments are presented in the Tab. ~\ref{tab:burgers2_mae}. The choice of importance distribution has significantly influenced the algorithm's performance, so that the correct structure of the equation is found in all runs. On the other hand, it can be noted that the sparse regression package (PySINDy) found the desired equation with a smaller error in coefficients.

\begin{table}[h!]
\centering
\caption{Coefficients MAE, Burger's equation. The green colour marks minimal obtained error, and 'n/a' marks run when the algorithm cannot restore the equation with given importance distribution}
\label{tab:burgers2_mae}
\begin{tabular}{|c|c|c|c|c|c|}
\hline
\begin{tabular}[c]{@{}c@{}}Basic\\ algorithm\end{tabular} &
  \begin{tabular}[c]{@{}c@{}}Fixed\\ initial\\ distr.\end{tabular} &
  \begin{tabular}[c]{@{}c@{}}Biased\\ initial\\ distr.\end{tabular} &
  \begin{tabular}[c]{@{}c@{}}Highly\\ biased\\ distr.\end{tabular} &
  \begin{tabular}[c]{@{}c@{}}Uniform\\ initial\\ distr.\end{tabular} &
  \multicolumn{1}{l|}{PySINDy} \\ \hline
\cellcolor[HTML]{C5E0B3}0.0048 & \cellcolor[HTML]{C5E0B3}0.0048 &
  \cellcolor[HTML]{C5E0B3}0.0048 & \cellcolor[HTML]{C5E0B3}0.0048 & 0.0049 & 0.0004 \\ \hline
\cellcolor[HTML]{C0C0C0}n/a &
  \cellcolor[HTML]{C5E0B3}0.0048 & 0.0049 &
  \cellcolor[HTML]{C5E0B3}0.0048 &
  \cellcolor[HTML]{C0C0C0}{\color[HTML]{000000} n/a} & 0.0004
   \\ \cline{1-6} 0.0049 &
  \cellcolor[HTML]{C5E0B3}0.0048 & \cellcolor[HTML]{C5E0B3}0.0048 &
  \cellcolor[HTML]{C5E0B3}0.0048 & \cellcolor[HTML]{C5E0B3}0.0048 & 0.0004
   \\ \cline{1-6}
\cellcolor[HTML]{C5E0B3}0.0048 & \cellcolor[HTML]{C5E0B3}0.0048 &
  \cellcolor[HTML]{C5E0B3}0.0048 & 0.0049 & 0.0049 & 0.0004
   \\ \cline{1-6} 0.0049 &
  \cellcolor[HTML]{C0C0C0}n/a & \cellcolor[HTML]{C5E0B3}0.0048 &
  \cellcolor[HTML]{C5E0B3}0.0048 & \cellcolor[HTML]{C5E0B3}0.0048 & 0.0004
   \\ \cline{1-6}
\cellcolor[HTML]{C5E0B3}0.0048 & \cellcolor[HTML]{C5E0B3}0.0048 &
  \cellcolor[HTML]{C5E0B3}0.0048 & \cellcolor[HTML]{C5E0B3}0.0048 &
  \cellcolor[HTML]{C5E0B3}0.0048 & 0.0004
   \\ \cline{1-6}
\cellcolor[HTML]{C0C0C0}n/a & \cellcolor[HTML]{C5E0B3}0.0048 &
  \cellcolor[HTML]{C5E0B3}0.0048 & \cellcolor[HTML]{C5E0B3}0.0048 &
  \cellcolor[HTML]{C5E0B3}0.0048 & 0.0004
   \\ \cline{1-6}
\cellcolor[HTML]{C5E0B3}0.0048 & 0.0049 &
  \cellcolor[HTML]{C5E0B3}0.0048 & \cellcolor[HTML]{C5E0B3}0.0048 &
  \cellcolor[HTML]{C5E0B3}0.0048 & 0.0004
   \\ \cline{1-6}
\cellcolor[HTML]{C5E0B3}0.0048 & \cellcolor[HTML]{C0C0C0}n/a &
  \cellcolor[HTML]{C5E0B3}0.0048 & \cellcolor[HTML]{C5E0B3}0.0048 &
  \cellcolor[HTML]{C5E0B3}0.0048 & 0.0004
   \\ \cline{1-6}
\cellcolor[HTML]{C5E0B3}0.0048 &
  \cellcolor[HTML]{C0C0C0}n/a & \cellcolor[HTML]{C5E0B3}0.0048 &
  \cellcolor[HTML]{C5E0B3}0.0048 & \cellcolor[HTML]{C0C0C0}n/a & 0.0004
   \\ \cline{1-6}
\end{tabular}
\end{table}

\begin{figure}[h!]
    \centering
    \includegraphics[width=0.9\linewidth]{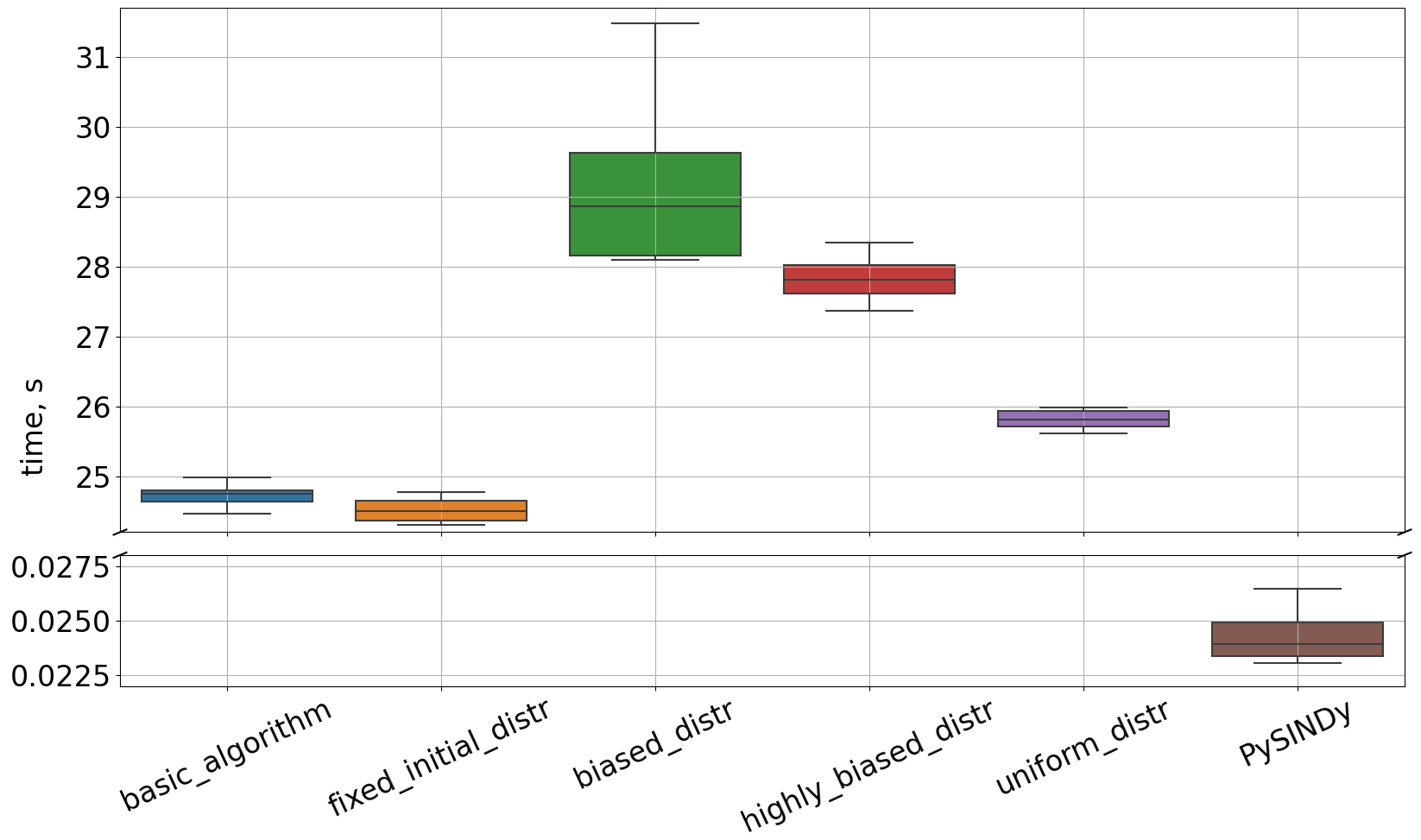}
    \caption{Algorithm running time, Burgers' equation from left to right: classical algorithm, modified algorithm with classical term importance distribution, moderate and significant Burger's term importance increase, uniform distribution.}
    \label{fig:time_burgers2}
\end{figure}

The time consumption is illustrated in Fig.~\ref{fig:time_burgers2}. With increasing term importance, the algorithm needs more time to find the solution. Compared with PySINDy, the algorithm is much slower and yet has more flexibility, making it possible to find, for example, an inhomogeneous Korteweg -- de Vries equation. Note that the PySINDy package is based on deterministic algorithms. Hence it has the same value of error for all ten runs.

\subsection{Comparison with PySINDy: Korteweg -- de Vries equation}

The solution \footnote{https://github.com/dynamicslab/pysindy} to the problem in Eq.~\ref{eq:KdV_homo} was provided by the authors \cite{desilva2020,Kaptanoglu2022} at the domain $(x,t) \in [-30,30] \times [0,20]$ using $ 512 \times 201$ discretization points.

\begin{equation}
\begin{array}{cc}
\frac{\partial u}{\partial t} + 6 u \frac{\partial u}{\partial x} + \frac{\partial^3 u}{\partial x^3} = 0 \\
(x,t) \in [-30,30] \times [0,20]
\end {array}
    \label{eq:KdV_homo}
\end{equation}

\begin{figure}[h!]
    \centering
    \includegraphics[width=0.9\linewidth]{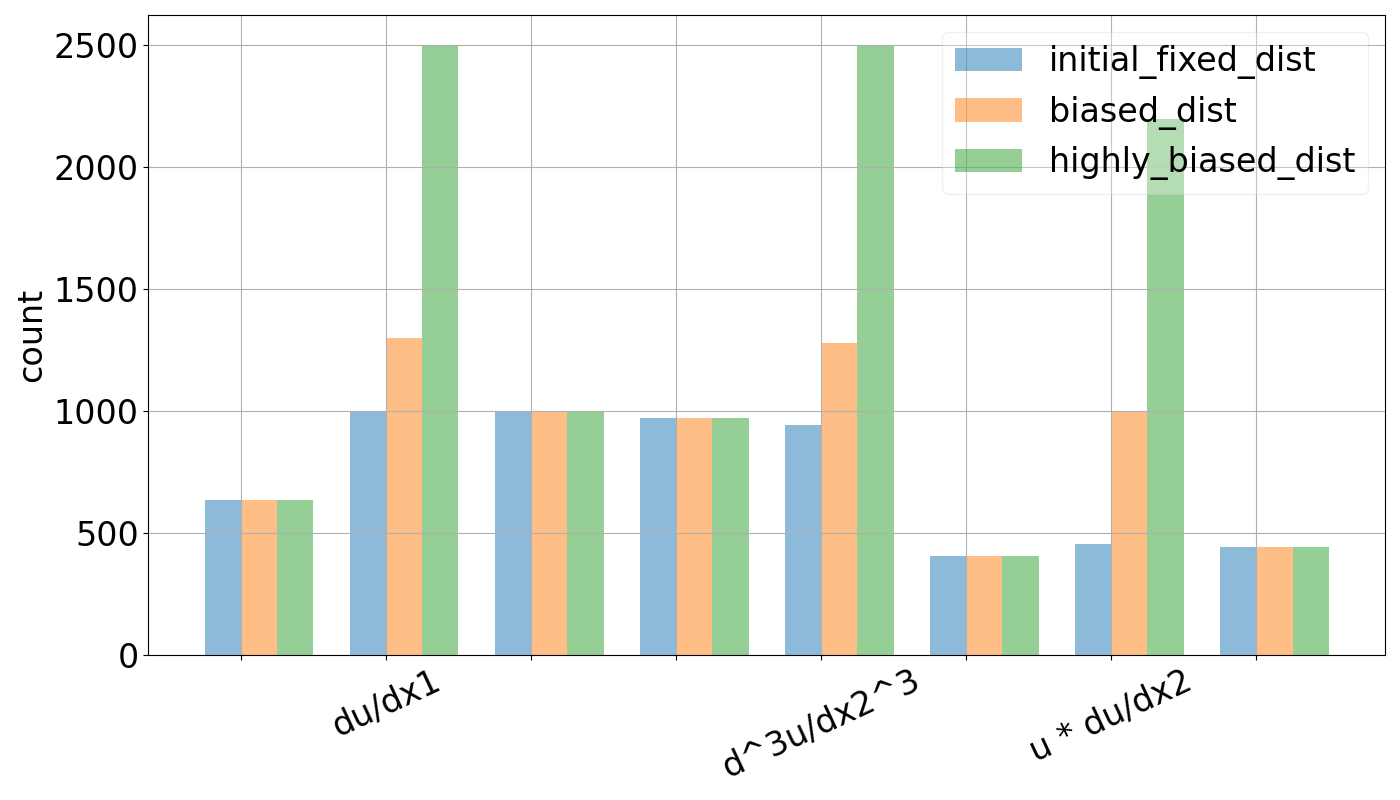}
    \caption{Importance distributions forms for cases classical algorithm (i, blue), the classical algorithm with a moderate increase (ii, orange) and significant increase (iii, green) of the Korteweg -- de Vries equation terms importance.}
    \label{fig:dists_kdv2}
\end{figure}

The three possible distributions, excluding the uniform distribution, formed in the same way as in the previous experiments are depicted in Fig.~\ref{fig:dists_kdv2}.

Experimental results are shown in Tab. ~\ref{tab:mae_kdv2}. The algorithm achieved the best performance with a slight increase in term importance - it was able to find the equation in every run and often with less error than other types of the algorithm. In this experiment, the PySINDy package has a higher error in the equation coefficients than all algorithms considered.

\begin{table}[h!]
\centering
\caption{Coefficients MAE, Korteweg -- de Vries equation. The green color marks the minimal obtained error, and 'n/a' marks run when the algorithm cannot restore the equation with given importance distribution.}
\label{tab:mae_kdv2}
\begin{tabular}{|c|c|c|c|c|c|}
\hline
\begin{tabular}[c]{@{}c@{}}Basic\\ algorithm\end{tabular} &
  \begin{tabular}[c]{@{}c@{}}Fixed\\ initial\\ distr.\end{tabular} &
  \begin{tabular}[c]{@{}c@{}}Biased\\ initial\\ distr.\end{tabular} &
  \begin{tabular}[c]{@{}c@{}}Highly\\ biased\\ distr.\end{tabular} &
  \begin{tabular}[c]{@{}c@{}}Uniform\\ initial\\ distr.\end{tabular} &
  \multicolumn{1}{c|}{PySINDy} \\ \hline
0.0039 & 0.0039 & \cellcolor[HTML]{C5E0B3}0.0031 & 0.0039 & 0.0039 & 0.0138 \\ \hline 
  0.0039 & \cellcolor[HTML]{C5E0B3}0.0031 & 0.0039 & 0.0039 & 0.0039 & 0.0138 
   \\ \cline{1-6}
\cellcolor[HTML]{C5E0B3}0.0031 & 0.0039 & \cellcolor[HTML]{C5E0B3}0.0031 &
  0.0039 & 0.0039 & 0.0138 
   \\ \cline{1-6}
\cellcolor[HTML]{C5E0B3}0.0031 & 0.0039 & 0.0039 & 0.0039 & 0.0039 & 0.0138 
   \\ \cline{1-6}
\cellcolor[HTML]{C5E0B3}0.0031 & 0.0039 & \cellcolor[HTML]{C5E0B3}0.0031 &
  0.0039 & 0.0039 & 0.0138 
   \\ \cline{1-6}
0.0039 & 0.0039 & \cellcolor[HTML]{C5E0B3}0.0031 &
  0.0039 & 0.0039 & 0.0138 
   \\ \cline{1-6}
\cellcolor[HTML]{C5E0B3}0.0031 & 0.0039 &
  \cellcolor[HTML]{C5E0B3}0.0031 & 0.0039 & 0.0039 & 0.0138 
   \\ \cline{1-6}
\cellcolor[HTML]{C0C0C0}n/a & 0.0039 &
  \cellcolor[HTML]{C5E0B3}0.0031 & 0.0039 & 0.0039 & 0.0138 
   \\ \cline{1-6}
0.0039 & 0.0039 & 0.0039 & 0.0039 & 0.0039 & 0.0138 
   \\ \cline{1-6}
\cellcolor[HTML]{C5E0B3}0.0031 & 0.0039 & \cellcolor[HTML]{C5E0B3}0.0031 
& \cellcolor[HTML]{C5E0B3}0.0031 & \cellcolor[HTML]{C5E0B3}0.0031 & 0.0138 
   \\ \cline{1-6}
\end{tabular}
\end{table}

\begin{figure}[h!]
    \centering
    \includegraphics[width=0.9\linewidth]{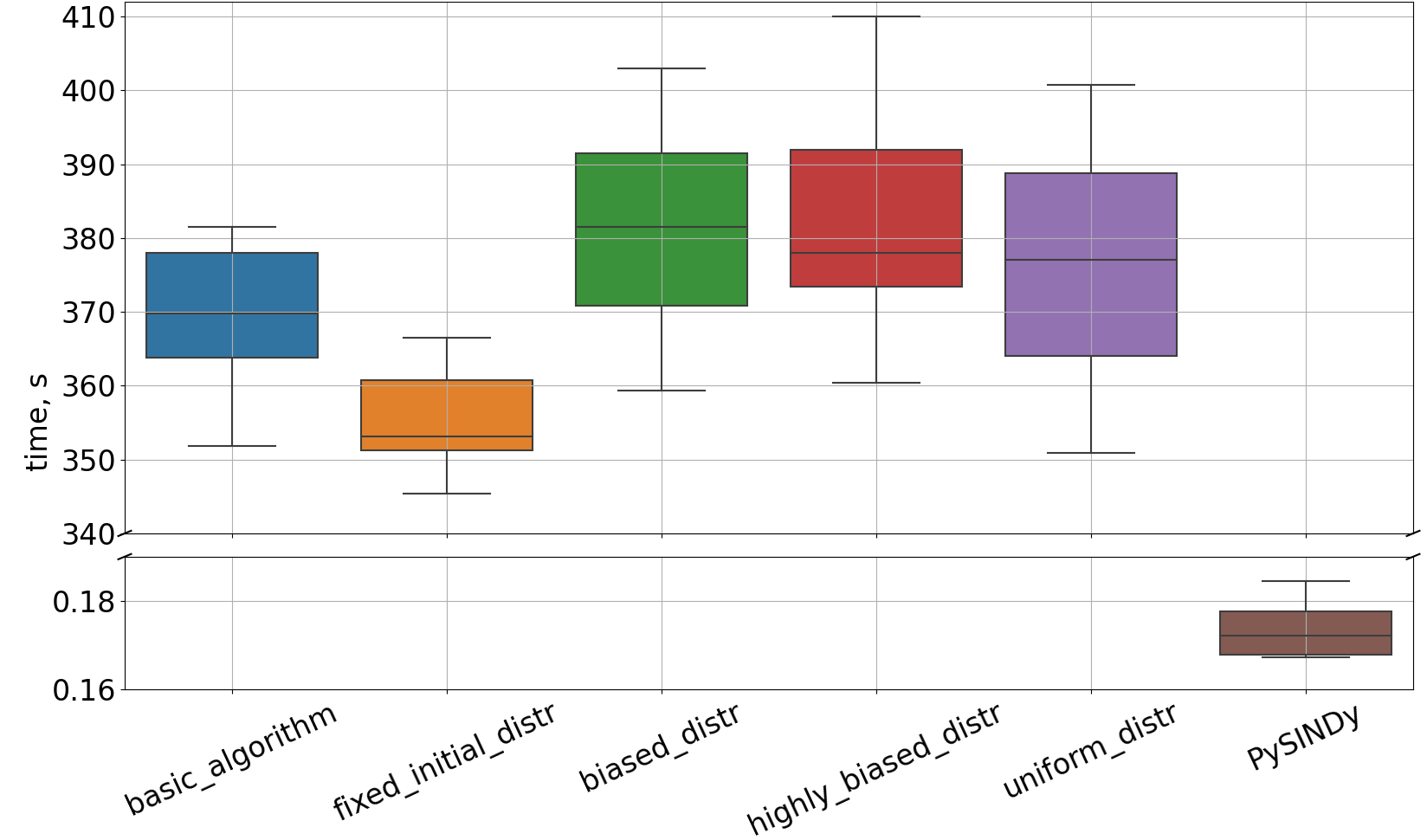}
    \caption{Algorithm running time, Korteweg -- de Vries equation from left to right: classical algorithm, the modified algorithm with classical term importance distribution, moderate and significant Korteweg -- de Vries equation term importance increase, uniform distribution.}
    \label{fig:time_kdv2}
\end{figure}

The time assessment required to find the equation is shown in Fig.~\ref{fig:time_kdv2}. An increase in time consumption can be observed for the algorithm with modified distributions. PySINDy package spends less time finding the correct equation. However, as mentioned previously, the package has some limitations.

To conclude the experiments, we note that the moderate term importance increase leads to better results of the equation structure discovery. It may lead to the optimization time increase in complex cases, but higher time consumption is reasonably compensated with the overall better structure of the found equation, which is reflected in Tab. ~\ref{tab:total_errors}.

\begin{table}[h!]
\centering
\caption{Characteristics of runs for each type of the algorithm, where the least error in coefficients was obtained (total number of runs is 50 for each column).}
\label{tab:total_errors}
\begin{tabular}{|c|c|c|c|c|c|}
\hline
Criteria &
  \begin{tabular}[c]{@{}c@{}}Basic\\ algo-\\ rithm\end{tabular} &
  \begin{tabular}[c]{@{}c@{}}Fixed\\ initial\\ distr.\end{tabular} &
  \begin{tabular}[c]{@{}c@{}}Biased\\ initial\\ distr.\end{tabular} &
  \begin{tabular}[c]{@{}c@{}}Highly\\ biased\\ distr.\end{tabular} &
  \begin{tabular}[c]{@{}c@{}}Uniform\\ initial\\ distr.\end{tabular} \\ \hline
\begin{tabular}[c]{@{}c@{}}Number of runs \\ with the least \\ obtained error\end{tabular} & 26 & 21 & 39 & 26 & 22 \\ \hline
\begin{tabular}[c]{@{}c@{}}Percentage of the\\ runs with minimal\\ error among all runs\end{tabular} &
  52\% & 42\% & 78\% & 52\% & 44\% \\ \hline
\end{tabular}
\end{table}

In conclusion, we make note of a non-trivial finding: while increasing the probabilities of terms present in an equation may produce results to a certain extent, this goes against common sense. The outcome of such an approach is the creation of a population primarily composed of ``long'' terms, which lacks the ability to generate simpler ones. Consequently, a higher probability should be assigned to choosing terms that are already present in the equation, but not to a dominant degree.

\section{Discussion and conclusion}
\label{sec:conclusion}

In the paper, we propose modifications to the classical crossover and mutation operators of the evolutionary equation discovery algorithm. We improve the algorithm's ability to converge towards a given equation more frequently without affecting the overall optimization time. 

Although our initial testing was limited to cases with correct data and derivatives, to fully evaluate our proposed approach, further testing with noised data is necessary. Nevertheless, our preliminary results indicate that the modified operators show promise and may result in an increase in the quality of the restored equation for cases that are closer to experimental data. 

Compared to the classical gradient-based sparse regression approach, our proposed evolutionary approach not only offers more flexibility, but also produces restored equation quality of at least the same level. In addition, it is capable of restoring an equation that might have been found incorrectly by gradient algorithms in complex cases. However, one downside to the evolutionary approach is that the optimization time is typically longer. Despite this, we observe that, by incorporating domain knowledge, the success rate of the algorithm can increase from an average of 52\% to 78\%. 

To enable the automatic incorporation of domain knowledge into the algorithm, one would need a tool capable of deducing the distribution of term importance from observational data. This could take the form of either classical knowledge-extracting tools from the literature or modern one-shot learning convolutional networks. Developing such knowledge extraction tools is an important aspect of our ongoing work.

\section{Code and data availability}

Code and data to reproduce the experimental results are openly available in the repository \url{https://github.com/ITMO-NSS-team/CEC_2023_knowledge}.

\section*{Acknowledgment}

This work was supported by the Analytical Center for the Government of the Russian Federation (IGK 000000D730321P5Q0002), agreement No. 70-2021-00141.

\bibliographystyle{IEEEbib}
\bibliography{references}

\end{document}